\documentclass[a4paper,10pt,twoside]{article}

\usepackage{clin}        
\usepackage{harvard}     
\usepackage{tabularx}
\usepackage{booktabs}
\usepackage{array}
\usepackage{url}
\usepackage{float}
\usepackage{placeins}
\usepackage{amsmath}
\usepackage{ragged2e}
\usepackage{xcolor}
\usepackage{enumitem}
\usepackage{graphicx}
\usepackage{refcount}
\definecolor{FemalePurple}{HTML}{87579A}
\definecolor{MaleOrange}{HTML}{CC6502}
\definecolor{ProSBlue}{HTML}{6C8EBF}
\definecolor{AntiSGreen}{HTML}{81B365}

\begin{document}

\title{Is \textit{She} Even Relevant? When BERT Ignores Explicit Gender Cues}

\author{Jonas Klein \email{j.b.klein@tilburguniversity.edu}\\
{\normalsize \bf Chiara Manna  }\email{c.manna@tilburguniversity.edu}\\
{\normalsize \bf Eva Vanmassenhove}\email{e.o.j.vanmassenhove@tilburguniversity.edu}
\AND \addr{Department of Computational Cognitive Science, Tilburg University, Warandelaan 2, 5037 AB Tilburg, Netherlands}}

\maketitle\thispagestyle{empty} 


\begin{abstract}

Gender bias in large language models has primarily been investigated for English, while languages with grammatical or morphological gender remain comparatively understudied. This paper investigates how and when gender information emerges in a Dutch BERT model trained from scratch, offering one of the first checkpoint-level analyses of bias formation in a Transformer architecture for a language combining overt morphological gender marking and generic forms. By extracting contextual embeddings throughout training, we construct dynamic gender subspaces using linear SVMs to trace when gender becomes linearly encoded and how this encoding evolves over time. Contextual embeddings are often assumed to integrate contextual cues robustly, allowing models to adjust the representation of a word depending on its more local usage. We therefore test whether explicit gender cues in controlled sentence templates (e.g., \textit{Zij} is een loodgieter (‘\textit{She} is a plumber’)) can override learned statistical associations (\textit{plumber} $\rightarrow$ \textit{male}). Our findings challenge this assumption: although gender becomes clearly linearly separable around epoch 20 and is distributed across multiple embedding dimensions, the model struggles to update its internal gender representation in light of explicit contextual cues in short sentence templates. Stereotypical gender–profession pairings are predicted far more accurately than anti-stereotypical ones, and generic forms in Dutch systematically default to a male interpretation, even when the context explicitly denotes a female referent. 
Together, our results seem to indicate that contextualization in the representations learned by our Dutch BERT model is not sufficiently dynamic along the probed gender direction: explicit gender cues in anti-stereotypical contexts are not reliably reflected in the resulting representations, resulting in persistent male-default behaviour. These findings underscore once more the challenges for contextual modeling and gender (bias) in multilingual NLP.

\end{abstract}

\newpage 

\section{Introduction} \label{sec:introduction}

If the model’s representation of a \textit{plumber} is consistently male-coded, what does that reveal about the kind of semantic and social knowledge it encodes, and how might this affect downstream task performance? While both humans and models pick up on statistical regularities from exposure \cite{caliskan2017semantics}, humans can override stereotypes through world knowledge, pragmatic reasoning and explicit contextual cues. Models like BERT \cite{BERT}, by contrast, rely strongly on distributional patterns in texts. When these patterns are uneven, the resulting internal representations inherit these asymmetries. This means that stereotypical associations, such as \textit{plumber} $\rightarrow$ \textit{male} or \textit{nurse} $\rightarrow$ \textit{female} can become persistent defaults in the model's embeddings space, even when contradicted by explicit contextual cues. Stereotypes become essentially hard-coded and at scale, these defaults raise concerns about how such biases could be amplified in downstream applications, and maybe reducing diversity and plurality in models' outputs \cite{vanmassenhove2025losing}.
Large Language Models (LLMs) have repeatedly been shown to reproduce and sometimes intensify such patterns \cite{StochasticParrots}, with consequences for downstream tasks such as translation \cite{vanmassenhove-etal-2018-getting,saunders-byrne-2020-reducing}, coreference resolution \cite{rudinger-etal-2018-gender,zhao-etal-2018-gender}, and text generation \cite{fleisig-etal-2023-fairprism,sheng-etal-2019-woman}. 

Zooming in on gender bias specifically, a large body of research has focused on English, while languages with grammatical and morphological gender have been underexplored, even though they introduce additional layers of complexity for analysing gender (bias) in Natural Language Processing (NLP). From a linguistic perspective, Dutch occupies an intermediate position between fully gendered systems and largely genderless ones \cite{gerritsen2002towards}, combining residual grammatical gender with ongoing processes of neutralisation \cite{mortelmans2008zij,boudewijn2023masculinegeneric}. In Dutch, for instance, many professional titles are asymmetrically marked: historically male, morphologically unmarked forms increasingly function as generic terms, while grammatically female forms explicitly refer to someone who identifies as a woman and are often morphologically derived. This asymmetry reflects a broader tension between neutralisation and feminisation in the Dutch system, where overt gender marking is reduced but not eliminated. Moreover, this dynamic is not uniform across the Dutch-speaking area: while neutralisation dominates in the Netherlands, Flemish usage retains more productive and socially accepted female forms \cite{boudewijn2023masculinegeneric}. These properties make Dutch an interesting language to investigate how contextual cues can interact with the statistically learned priors in language models.


To analyze and detect biases in word embeddings, projection-based methods have been widely used \cite{BolukbasiCZSK16a,ravfogel-etal-2020-null,vanderwal2022birthbiascasestudy}. Our approach is most similar to the one described in \citeasnoun{vanderwal2022birthbiascasestudy}, who traced the evolution of gender bias in an English Long Short-Term Memory (LSTM) model deriving dynamic gender subspaces using linear Support Vector Machines (SVMs). We extend their approach to a \textit{Dutch} BERT model trained entirely from scratch, allowing us to assess whether the patterns observed by \citeasnoun{vanderwal2022birthbiascasestudy} generalize to Transformer-based architectures \cite{Vaswanietal}, which underpin most modern NLP systems. By saving model checkpoints throughout pretraining, we track whether and how gender information becomes linearly separable over time. We then project profession terms within controlled sentence templates such as \textit{\textbf{Zij} is een l\textbf{oodgieter}} (``\textbf{She} is a \textbf{plumber}'') onto the learned gender subspace to test whether explicit gender cues are sufficient to override prior gender associations in the model’s representations or whether the representations continue to reflect stereotypical associations. We measure accuracy by checking whether each projected attribute embedding aligns with the gender specified by the sentence context rather than with the model's default gender associations.

Specifically, with our experiments we investigate how a Dutch BERT model encodes gender information during training, and to what extent its final profession representations systematically align with one side of a learned gender subspace derived from the training data.
A central question is whether explicit contextual cues e.g., \textit{zij} (`she') or \textit{hij} (`he'), are sufficient to override stereotypical priors and shift a profession embedding toward the contextually appropriate gender. To address this, we examine: (i) when gender becomes linearly separable; (ii) how it is distributed across embedding dimensions; (iii) the extent to which the representations align more strongly with stereotypical than with anti-stereotypical gender–profession pairings; and (iv) how morphological gender marking influences the degree to which contextual cues are integrated into the final representation.

The main contributions of this work align with these questions: (i) we extend prior analyses of bias formation in recurrent architectures \cite{vanderwal2022birthbiascasestudy} to a Transformer-based model, providing the first controlled, checkpoint-level investigation of how gender information emerges in contextual embeddings; (ii) we adapt projection-based subspace analysis to a contextual setting, combining it with controlled sentence templates to test how explicit gender cues interact with learned gender connotations; (iii) using Dutch, a language in flux between gendered and degendered usage (see Section \ref{sec:Dutch}), we show that even strong, explicitly female cues (e.g. female pronouns indicate a female referent) can be overridden by stereotypical male defaults; and finally, (iv) we demonstrate that contextual gender cues are frequently overshadowed by internal priors. Together, these findings establish a mechanistic baseline for understanding how structural gender bias forms and stabilizes in Transformer-based language models, and offer insight into processes that persist, though less transparently, in modern large-scale architectures.

\section{Related Work} \label{sec:relatedwork}
Although LLMs may appear to produce meaningful and coherent language, they operate by combining linguistic forms according to statistical patterns learned from large text corpora~\cite{mccoy2023embers}. They do not interpret or reason about meaning, which has led to their characterization as “stochastic parrots” \cite{StochasticParrots}. They therefore tend to reproduce and in some cases amplify biases present in training corpora \cite{vanmassenhove2019lost,vanmassenhove2021machine}. This behaviour poses risks, especially given the widespread use of LLMs for language generation and their integration into many downstream applications (e.g. Machine Translation \cite{saunders-byrne-2020-reducing}, hate speech recognition \cite{sap-etal-2019-risk}, text generation \cite{fleisig-etal-2023-fairprism,sheng-etal-2019-woman}, and many others.)\\

\paragraph{Bias in Static Word Embeddings}
Research into biases in word embeddings started early on with work on static word embeddings. One of the first systematic studies is by \citeasnoun{BolukbasiCZSK16a}, who showed that word2vec embeddings capture gender stereotypes such as “man is to computer programmer as woman is to homemaker.” They identify a gender direction in the embedding space by combining multiple gendered directions, such as she–he and woman–man. This is achieved by applying Principal Component Analysis (PCA), to vectors obtained by subtracting the embeddings of paired gendered words, to extract the primary axis that encodes gender. The bias of a word is then measured by projecting its embedding onto this gender direction.

Shortly after, \citeasnoun{caliskan2017semantics} propose a different approach to measuring bias in static word embeddings. Rather than explicitly modeling a gender subspace, they use a permutation test to illustrate and quantify bias through statistical associations. Their method considers two groups of occupation-related terms (e.g., programmer, engineer versus nurse, librarian) and examines how strongly they are associated with gendered terms, drawing on methodology from the Implicit Association Test (IAT) literature. The null hypothesis states that there is no difference in how closely the two sets of target words are associated with the two sets of attribute words. Across all tested scenarios, they reject the null hypothesis, providing evidence for the presence of bias in static word embeddings.

More recent work has revisited the subspace-based perspective. \citeasnoun{ravfogel-etal-2020-null} argue that the approach of identifying a single gender direction has a significant limitation: gender information is in fact distributed across hundreds of directions in the embedding space, while earlier methods rely on an intuitive selection of only a few. To address this, they propose Iterative Nullspace Projection (INLP), which trains linear classifiers to predict protected attributes Z (e.g., gender) from vector representations X (e.g., embeddings) and iteratively projects X onto the classifiers’ nullspaces to remove linearly encoded bias. This method is adopted by \citeasnoun{vanderwal2022birthbiascasestudy}, who use a Support Vector Machine (SVM) to determine the optimal linear decision boundary between clearly female and male words, with the orthogonal axis acting as the main gender subspace. Each input embedding can then be scalarly projected onto this subspace to determine its gender bias.

\paragraph{Bias in Contextual Models} Biases are not limited to static word embeddings; they also appear in contextual models, like ELMo \cite{peters-etal-2018-ELMo} and Transformer models like BERT \cite{BERT}. One early attempt to systematically measure such biases in contextual embeddings is SEAT \cite{may-etal-2019-measuring}, which extended traditional bias detection methods to sentence-level representations. To combine outputs into a fixed-sized vector, they employ pooling when necessary. \citeasnoun{Tan&Celis}, in turn, suggest evaluating bias at the level of contextual words, so looking at the embedding of the word of interest alone. This prevents confounding contextual effects at the sentence level, which might mask bias.

Gender bias in contextual models has also been studied through coreference resolution systems \cite{zhao-etal-2018-gender,rudinger-etal-2018-gender}. The goal of coreference resolution is to identify expressions (mentions) that refer to the same real-world entity. \citeasnoun{zhao-etal-2018-gender} created the WinoBias dataset, where sentences are created with a gendered pronoun and both a profession stereotypically carried out by that gender and a profession not stereotypically carried out by that gender. To pass the test, a system must be equally capable of making accurate coreference predictions in pro-stereotypical and anti-stereotypical contexts \cite{zhao-etal-2018-gender}. \citeasnoun{rudinger-etal-2018-gender} did a similar test, with the WinoGender dataset. Both studies demonstrate that coreference systems more frequently resolve pronouns following gender stereotypes rather than counter to them.

Another common approach to measuring contextual bias, without having to rely on coreference systems, is with the use of sentence templates \cite{bartl-etal-2020-unmasking}. This involves the use of pro-stereotypical and anti-stereotypical sentence pairs, assessing the difference in the model's assigned probabilities \cite{kurita-etal-2019-measuring,nangia-etal-2020-crows,nadeem-etal-2021-stereoset}. CrowS-Pairs \cite{nangia-etal-2020-crows} specifically includes pairs of sentences where one sentence is more stereotypical than the other, evaluating if models assign higher likelihoods to stereotypical contexts. \citeasnoun{nangia-etal-2020-crows} used sentence-level pseudo-log-likelihood scores. This score is computed by systematically masking each token in a sentence except the modified tokens, and summing the conditional probabilities of predicting these masked tokens given the context \cite{salazar-etal-2020-masked}. This allows for comparison between these summed scores across stereotypical and anti-stereotypical sentences. \citeasnoun{nangia-etal-2020-crows} found that sentences aligned with common societal stereotypes received consistently higher likelihood scores on various biases, including race, gender, religion, and socioeconomic status. Similarly, StereoSet \cite{nadeem-etal-2021-stereoset} presents contextually paired sentences where again the likelihood of stereotypical completions is contrasted with anti-stereotypical ones. \citeasnoun{kurita-etal-2019-measuring} came up with the log probability bias score by comparing how much more likely BERT is to associate one target (e.g., `he') over another (e.g., `she') with a given attribute (e.g., `programmer'), using sentence templates like `[TARGET] is a [ATTRIBUTE]', and normalizing this association by subtracting the model's prior bias estimated from a doubly masked template like `[MASK] is a [MASK]' \cite{kurita-etal-2019-measuring}.

Few attempts have been made to extend subspace projection techniques, originally developed for static word embeddings \cite{BolukbasiCZSK16a,ravfogel-etal-2020-null,vanderwal2022birthbiascasestudy}, to contextual embeddings. An exception is the work of \citeasnoun{kaneko-bollegala-2021-debiasing}, who proposed one of the first methods for projecting away bias in contextualised word embeddings such as BERT and RoBERTa~\cite{liu2019roberta}. Specifically, they extract contextual embeddings of gendered attribute words from an external corpus and use these to define a gender direction at each layer. They then fine-tune the model with a loss function that penalises alignment between target word embeddings (e.g., profession terms) and the gender direction, while simultaneously preserving semantic content through a regularisation term. Relatedly, \citeasnoun{liang-etal-2020-monolingual} apply DensRay to debias contextualized representations in BERT, computing a gender direction for each layer and probing individual attention heads. They find that debiasing a single attention head has minimal effect, while upper layers (7–10) show the strongest debiasing impact, suggesting that gender information is not localized but distributed across heads and layers.

While prior studies have applied sentence templates to probe stereotype alignment \cite{kurita-etal-2019-measuring,nangia-etal-2020-crows} and constructed contextual embedding subspaces to examine gender bias \cite{kaneko-bollegala-2021-debiasing,liang-etal-2020-monolingual}, few have combined these methods to directly project structured sentence contexts into learned subspaces. By intergrating these approaches, we can evaluate the influence of contextual cues on the model’s internal gender representations, something not measurable by template scoring or subspace construction in isolation.

Inspired by these approaches, we construct a contextual gender subspace by extracting embeddings of gendered words from diverse, naturally occurring contexts in the Dutch SoNaR corpus \cite{oostdijk2013construction}. Following the projection-based strategy of \citeasnoun{vanderwal2022birthbiascasestudy}, we train a linear SVM on thousands of contextual embeddings to define a dynamic gender axis. We then use this subspace to project embeddings of profession terms extracted from controlled sentence templates, allowing us to trace how gender cues in context (e.g., subject pronouns or morphological markers) interact with encoded bias (see Section~\ref{sec:methodology}).\\

\paragraph{Multilingual Bias and Grammatical Gender} Several studies have noted that gender bias research in NLP is heavily skewed toward English \cite{zhou-etal-2019-examining,gonen-etal-2019-grammatical,savoldi2025decade}. This presents a challenge, as methods developed for detecting gender bias in English often fail to generalize to grammatically gendered languages, where grammatical structures can obscure or distort the meaning of words \cite{bartl-etal-2020-unmasking}. In several natural languages, grammatical gender is also assigned to inanimate nouns, influencing inanimate noun word representations, making nouns of the same gender more similar to one another than nouns of different genders, even though they may be semantically different \cite{gonen-etal-2019-grammatical}. \citeasnoun{gonen-etal-2019-grammatical} compared word embeddings of inanimate noun pairs in German and Italian. They grouped these pairs based on grammatical gender (same vs. different) and measured their average cosine similarities. To reveal the bias introduced by gender agreement, they used English translations of the same pairs, where no grammatical gender exists, as a control. They found that same-gender noun pairs were significantly closer in embedding space in gendered languages compared to English. \citeasnoun{chavez-mulsa-spanakis-2020-evaluating} also state that little effort has been done to debias contextualized embeddings for other languages that have features that make it impossible to just import the original techniques, such as Dutch, where `zij' (\textsc{3sg.f}) means both `she' (\textsc{3sg.f}) and `they' (\textsc{3pl}) in English \cite{chavez-mulsa-spanakis-2020-evaluating}.

\citeasnoun{zhou-etal-2019-examining} tried to tackle these challenges by separating semantic gender from grammatical gender through the construction of two gender directions in the embedding space; a semantic and a grammatical gender direction. They determined grammatical gender by a collection of (grammatically) male and female nouns (such as water and table), and semantic gender by a collection of gender definition words (such as man and woman). \citeasnoun{zhou-etal-2019-examining} conclude that asymmetric projections of occupation terms on the semantic gender direction occur, whereas symmetric projections on the grammatical gender direction are symmetric with regard to the origin point. Female occupation words lean more towards the female side than male occupation words do towards the male side along the semantic gender direction, which demonstrates how the embeddings include different information for the two genders.

\citeasnoun{neveol-etal-2022-french} tackle a different problem related to the focus on English. While a lot of forms of bias are relevant in different regions around the world, some of the biases captured by CrowS-Pairs, such as those against African Americans, are specific to the societal context of the United States \cite{neveol-etal-2022-french}. Therefore, \citeasnoun{neveol-etal-2022-french} incorporated stereotypes specific to French culture and language. They argue that these culturally localized biases are often absent from datasets that are translated from English, since those do not capture biases unique to the French culture/language.
\citeasnoun{fort-etal-2024-stereotypical} continued with CrowS-Pairs by translating the pairs into seven additional languages, including Dutch, German, Arabic, and simplified Chinese. Their effort to scale bias evaluation across nine languages and cultural contexts shows how deeply connected linguistic and cultural factors are. Some stereotypes simply could not be translated. Moreover, they showed constructing minimal pairs in inflectional languages with grammatical gender can be tricky. These findings reinforce the point that bias evaluation methods developed for English do not transfer easily to other languages.


Dutch, though less grammatically gendered than languages like Spanish or German, still presents significant challenges. \citeasnoun{chavez-mulsa-spanakis-2020-evaluating} hypothesize that the mitigation step covers a smaller gender subspace than English and that the bias is reduced less as a result of the widespread use of the pronouns \textit{he} and \textit{she} in English, which cannot be directly transferred to Dutch due to pronoun ambiguity, as `zij' (\textsc{3sg.f}) means both `she' (\textsc{3sg.f}) and `they' (\textsc{3pl}).
Moreover, \citeasnoun{bartl-etal-2020-unmasking} show that in grammatically gendered languages such as German, measured gender bias is largely driven by morphological gender marking rather than social stereotypes. Female profession forms, which are marked (e.g. via the suffix -in added to an unmarked male base), are consistently associated more strongly with female person terms across all professions, including ones with approximately equal participation across genders. \citeasnoun{bartl-etal-2020-unmasking} therefore argue that standard bias detection methods, used for English, fail to isolate social gender bias in gender-marking languages, as grammatical gender often confounds the measurement of semantic or stereotypical associations. Dutch, where gender distinctions are also partially encoded morphologically, likely faces similar complications.


\section{A Note on Gender in Dutch}\label{sec:Dutch}
The behaviour of gender representations in our model cannot be interpreted independently from the linguistic system it is trained on. Dutch occupies a transitional position between grammatically/morphologically gendered and genderless languages, combining morphological gender with ongoing tendencies toward neutralisation. These structural properties directly shape how gender is encoded, marked, and interpreted in profession terms, both in human language use and in our experiments. This interaction between linguistic structure and language models has also been noted in applied contexts such as Machine Translation, where gender biases are reflected and sometimes even amplified in the outputs \cite{vanmassenhove-etal-2018-getting,steurs2021hoe}. For this reason, we first outline some aspects of Dutch gender morphology that are relevant to our research.

In terms of grammatical gender, Dutch occupies an intermediate position between a fully gendered system such as German and a largely genderless system such as English \cite{gerritsen2002towards}. Gender bias in Dutch has been discussed at least since the second feminist movement, when writers such as Annie Romein-Verschoor drew attention to the asymmetry of gender-specific profession terms and the invisibility of women under male generics \cite{romein1975over}. The contemporary Dutch system reflects several decades of language-internal change. As described by \citeasnoun{mortelmans2008zij}, grammatical gender distinctions in Dutch have become less salient: the traditional opposition between male and female nouns is disappearing, especially in the Netherlands. This reduction in overt gender marking facilitates the increasing trend toward \textit{neutralisation} \cite{boudewijn2023masculinegeneric}. Neutralisation as opposed to differentiation in the context of profession nouns, refers to a preference for morphologically unmarked (often historically male) profession terms instead of marked female ones. Because the language system no longer transparently encodes a male-female distinction, the (albeit historically grammatically male) profession nouns (e.g. \textit{loodgieter} (`plumber'), \textit{wetenschapper} (`scientist')) serve as generic terms.

This neutralisation, however, is far from complete in actual usage. Feminisation, using explicitly female profession forms derived by suffixation (e.g.\ \textit{-es}, \textit{-in}, \textit{-e}, \textit{-ster}) still occurs, particularly for occupations strongly associated with women. Forms such as \textit{lerares} (‘female teacher’) remain common, though even in these cases the unmarked (historically male) form (\textit{leraar}) is also used to refer to all genders. Many female forms, especially those involving older suffixes (e.g.\ \textit{directrice} vs.\ \textit{directeur} ‘female director’ vs.\ ‘director’; \textit{secretaresse} vs.\ \textit{secretaris} ‘female secretary’ vs.\ ‘secretary’) furthermore carry diminutive or even pejorative connotations, which often discourages their use and can strengthen the shift toward neutralisation. For high-prestige professions, speakers tend to avoid feminisation (e.g.\ \textit{advocaat} ‘lawyer’, \textit{kok} ‘chef’, \textit{journalist} ‘journalist’) although female forms exist \cite{boudewijn2023masculinegeneric}. In short, there is a complex tension: consistently differentiating between male and female profession terms seems to run counter to ongoing language change towards degendering, yet full neutralisation does not eliminate male-biased interpretations~\cite{vervecken2013changing,vervecken2015yes}. A further distinction can be made between morphologically unmarked profession nouns that function as generic forms (e.g.\ \textit{leraar} `teacher') and epicene (inherently gender-neutral) alternatives (e.g.\ \textit{leerkracht} `teacher'). It should be noted that these epicene forms are not explored in our study. In practice, the number of profession nouns in Dutch that are clearly lexically gender-neutral in this sense is relatively limited, and many commonly used role nouns are morphologically unmarked forms with a historical male origin.

\begin{figure}[htbp]
  \centering
  \includegraphics[width=0.35\textwidth]{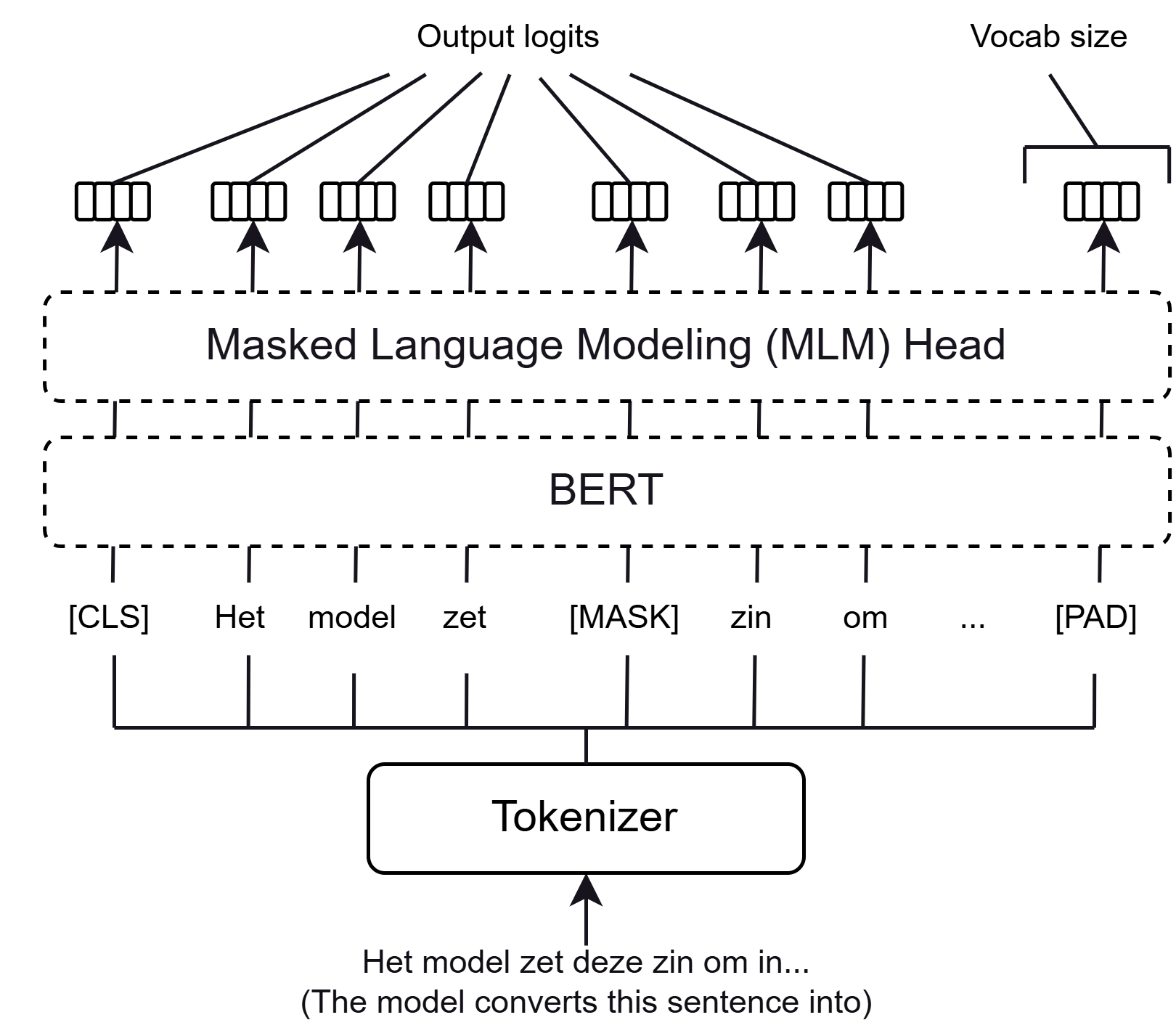}
  \caption{Architectural overview of the Dutch BERT model. The model processes input text from bottom to top: the Tokenizer converts raw sentences into token sequences, which are then fed into a 12-layer BERT encoder comprising self-attention and feed-forward sublayers. The output embeddings are passed through the MLM (Masked Language Modeling) head, which predicts masked tokens during pretraining. Contextual embeddings are extracted from the final hidden layer for bias analysis.}
  \label{fig:bert_architecture}
\end{figure}

It is furthermore worth mentioning that this dynamic does not unfold uniformly across the Dutch-speaking region. \citeasnoun{boudewijn2023masculinegeneric}’s corpus analysis demonstrates a regional split: in Flanders, female profession forms (\textit{lerares}, \textit{advocate}, \textit{chirurge}, \textit{studente}) remain more productive and socially acceptable. Flemish speakers not only use female forms more frequently, but also judge them as less marked and more appropriate than speakers in the Netherlands. In the Netherlands, by contrast, neutralisation dominates: unmarked and historically male profession terms are preferred in most contexts, and female forms are increasingly avoided or perceived as outdated. {At the same time, there are some indications that the underlying grammatical gender system in Belgian Dutch is also starting to undergo a gradual chance~\cite{van2026gender}. \citeasnoun{van2026gender}, for instance, discusses how the gender system in Belgian Dutch remains more robust due to dialectal support, but is increasingly subject to uncertainty amoung younger speakers who are not as familiar with dialects. This might suggest that, although female marking is comparatively productive in present-day Flemish, the conditions that support it may be weakening over time.\\

\section{Method} \label{sec:methodology}
Our methodological approach builds on and connects several strands of the prior work. Whereas earlier studies have examined bias either through static gender subspaces \cite{BolukbasiCZSK16a,ravfogel-etal-2020-null} or through sentence-level stereotype tests in contextual models \cite{kurita-etal-2019-measuring}, our work integrates the two approaches. From this perspective, we define gender bias not merely as a fixed association, but as the ``tendency of models to default to learned statistical associations rather than systematically relying on contextual information for gender disambiguation'' \cite{manna2025payingattentionherinvestigating}. Under this understanding, we analyze how gender bias emerges and evolves throughout the training of a Dutch BERT model by tracking how gender representations form and assessing whether they can be overridden by contextual cues. To this end, we (i) train BERT from scratch on a controlled Dutch corpus (see Figure~\ref{fig:bert_architecture} for an overview of the model architecture), 
(ii) derive gender subspaces at multiple training checkpoints, and 
(iii) project our target embeddings onto these subspaces. Together, these steps allow us to determine when gender becomes linearly separable, how gender information is distributed across the embedding space, and whether contextual cues can override the model’s learned gender priors. To further illustrate this process, we present an overview of the methodological pipeline of the gender subspace construction and the bias evaluation strategy in Figure ~\ref{fig:data_pipeline_vertical}.

\begin{figure}[htbp]
    \centering
    \includegraphics[width=\linewidth]{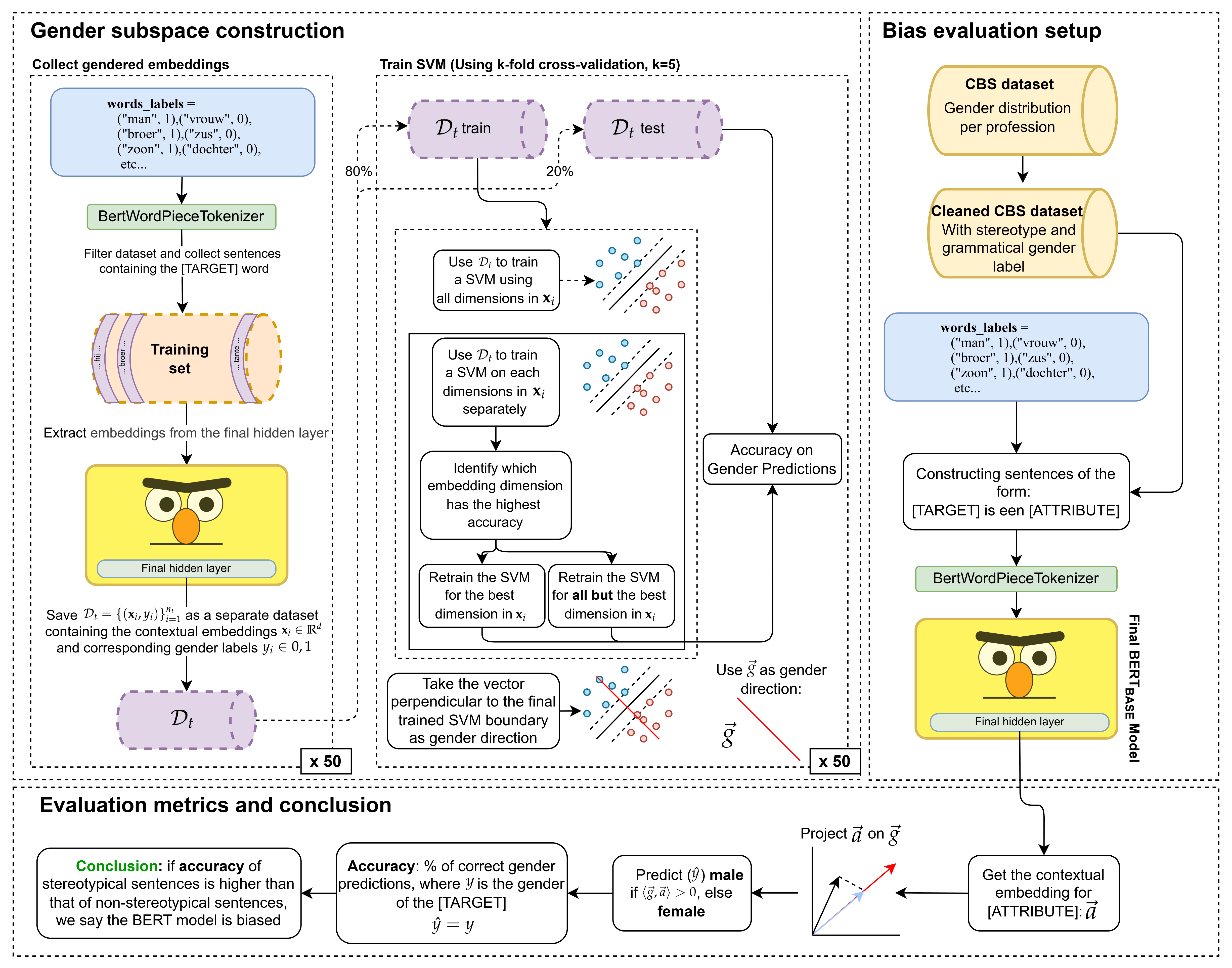}
    \caption{
    Methodological pipeline of the gender subspace construction and the bias evaluation. At each checkpoint, a gender subspace is constructed using SVMs trained on contextual embeddings of gendered words. Bias is evaluated by projecting profession terms from controlled sentences onto these subspaces, measuring accuracy. \textit{Note: Colors do not reflect the stereotype-related color scheme defined in section \ref{sec:methodology}.}
    }
    \label{fig:data_pipeline_vertical}
\end{figure} 

\subsection{Data} \label{sec:corpus}
Following \citeasnoun{vanderwal2022birthbiascasestudy}, we train BERT from scratch to retain full control over the training data and to enable the systematic extraction of intermediate model checkpoints during training. We use the books subset of the fully accessible Dutch SoNaR-500 corpus \cite{oostdijk2013construction} ($\approx 26$M words, 508 books), which provides us with a diverse and stylistically varied training corpus of manageable size under our computational constraints. We prepare the corpus by constructing a 30k-token vocabulary, with a minimum frequency threshold of 3, to capture the most frequent tokens while limiting the number of \texttt{[UNK]} placeholders. We further limit our character set to 1,000 unique characters, and include standard special tokens (\texttt{[PAD]}, \texttt{[UNK]}, \texttt{[CLS]}, \texttt{[SEP]}, \texttt{[MASK]}). Sentences are then tokenised using the \texttt{nltk} tokeniser \cite{bird2009natural} and chunked into sequences of at most 512 tokens (BERT's max input length). Finally, an overlap (stride) of 384 tokens is applied to preserve contextual continuity across chunk boundaries. The processed data is partitioned by token count into training (80\%), validation (10\%), and test (10\%) sets. We make sure to avoid splitting any books across these sets to prevent data leakage. Moreover, we intentionally preserve the naturally unbalanced gender distribution of the corpus to observe how gendered patterns may emerge from raw, unbalanced input. This setup ensures that any bias we observe originates directly from the SoNaR corpus rather than from unknown (often English-dominant) pretraining data.

Furthermore, the decision to train a BERT model from scratch is motivated by the nature of our research goal: rather than evaluating model outputs, we aim to trace \textit{when} and \textit{how} gender representations emerge during training. This requires full control over the training process and access to intermediate training checkpoints, conditions that are not easily met with larger (decoder-only) models under realistic computational constraints. At the same time, although recent advancements in NLP have been largely driven by decoder-only architectures, many important tasks, such as information retrieval and classification, still rely on general-purpose vector representations obtained from bidirectional encoder models. As a matter of fact, encoder-only models remain widely deployed in practice and are experiencing renewed research interest, as evidenced by the recent release of ModernBERT \cite{warner2024modernbert}, EuroBERT \cite{boizard2025eurobert}, and NeoBERT \cite{lebreton2025neobert}, which apply modern post-training techniques to encoder-only architectures.

\subsection{Gender Subspace Construction} \label{sec:gender_subspace}
To assess how the BERT model encodes gender information, we train a binary SVM classifier on the contextual embeddings of gendered Dutch words to derive dynamic gender subspaces across training checkpoints. For this purpose, we use a curated list of 60 gendered word pairs adapted from the English list in \citeasnoun{vanderwal2022birthbiascasestudy} (see Appendix, Table~\ref{tab:gender_pairs}), excluding ambiguous forms (e.g., \textit{zij}, \textit{haar}) to avoid polysemous interference.\footnote{“zij” can mean both “she” and “they,” and “haar” can mean both “her” and “hair.”} Although the SoNaR corpus provides POS tags that can distinguish between meanings, we do not rely on them. In models such as BERT, different meanings of a word share the same base embeddings and are only separated through context (in deeper layers). This means that even POS-filtered examples can still contain variation unrelated to gender. Because our goal is to learn a clear gender direction, such variation would introduce noise and make the results less reliable. We therefore include only unambiguous forms. \\ At each training checkpoint $t$, we extract the final hidden states for all target word occurrences in our training corpus. To prevent label imbalance, each target word is downsampled to 200 occurrences (see Appendix, Table \ref{tab:embedding_pairs}). Since words are often split into multiple subword tokens, we average the final hidden states of all subword pieces to obtain a single vector per word occurrence \cite{hewitt-manning-2019-structural}.  This results in a dataset $\mathcal{D}_t = \{(\mathbf{x}_i, y_i)\}_{i=1}^{n_t}$, where $\mathbf{x}_i \in \mathbb{R}^{768}$ is the contextual embedding associated with a word occurrence and $y_i \in \{0,1\}$ denotes its grammatical gender (0 = female, 1 = male).

Using  $\mathcal{D}_t$, we train a linear Support Vector Machine (SVM) \cite{SVMs} at each checkpoint $t$ with a 5-fold group-aware cross-validation (\texttt{GroupKFold}), grouping embeddings by the associated target word to ensure that all occurrences of a given word appear exclusively in either the train or test split. The trained classifier returns a weight vector $\mathbf{w}_t \in \mathbb{R}^d$ and intercept $b_t \in \mathbb{R}$ such that the decision boundary satisfies:

\begin{equation}
f_t(\mathbf{x}) = \text{sign}(\langle \mathbf{w}_t, \mathbf{x} \rangle + b_t)
\label{eq:svm}
\end{equation}
\\ where $\mathrm{sign}(\cdot)$ returns $\mathbf{+1}$ if its argument is positive and $\mathbf{-1}$ otherwise, assigning each embedding to either the male ($\mathbf{+1}$) or female ($\mathbf{-1}$) class (we treat zero as belonging to the negative class). Following \citeasnoun{vanderwal2022birthbiascasestudy}, we interpret the weight vector $ \mathbf{w}_t $ as the model's \textbf{gender direction} at checkpoint $t$. The scalar projection of any embedding $\mathbf{x}$ onto this direction provides us with a gender score: 
\[
\text{bias}_{\text{SVM}}(\mathbf{x}) = \langle \mathbf{w}_t, \mathbf{x} \rangle.
\] 
Positive values indicate alignment with the male class and negative values with the female class. 

By applying this procedure independently at each checkpoint, we derive a series of gender subspaces $\{{\mathbf{w}}_t\}_{t=1}^{T}$, which allow us to track how linearly separable gender becomes over the course of training. Following \citeasnoun{vanderwal2022birthbiascasestudy}, we quantify the strength of the learned gender direction at each checkpoint by computing the SVM's classification accuracy on $\mathcal{D}_t$. 

Furthermore, we assess if and where gender information is localized within the embedding space, by repeating the classification using a single embedding dimension at a time. Concretely, for each dimension $j \in \{1,\dots,768\}$ and checkpoint $t$, we construct a one-dimensional dataset 
$\mathcal{D}_t^{(j)} = \{(x_{i}^{(j)}, y_i)\}$, where $x_{i}^{(j)}$ is the $j$-th coordinate of $\mathbf{x}_i$, and train a separate linear SVM on $\mathcal{D}_t^{(j)}$. This allows us to assess which individual dimensions carry the most gender-relevant information. 

Finally, to characterize asymmetries between the male and female classes, we analyze per-gender recall patterns across embedding dimensions. Using the same SVM classifiers as before, trained on individual dimensions, we compute recall scores for the female and male classes. We then apply K-Means clustering to identify 30 groups of dimensions with similar recall. 

\subsection{Bias Evaluation Setup} \label{sec:bias_setup}

Once the gender subspaces are defined, we evaluate the BERT model’s gender bias and examine the extent to which contextual cues override its learned gender priors. For this purpose, we follow the controlled sentence template approach introduced by \citeasnoun{zhao-etal-2018-gender} and later extended by \citeasnoun{kurita-etal-2019-measuring}:

\begin{center}
\textit{[TARGET] is een [ATTRIBUTE]} \\
{\scriptsize \textit{([TARGET] is a [ATTRIBUTE])}}
\end{center} 

In this setup, the \textbf{attribute word} is a \textbf{profession term} (e.g.\ \textit{arts} (‘doctor’), \textit{verpleger} (‘nurse’), \textit{verpleeg\textbf{\textcolor{FemalePurple}{ster}}} (‘female nurse’)). These attribute words can be either unmarked forms (albeit historically male, in contemporary Dutch they function as generic terms), or explicitly female forms derived through suffixation (e.g.\ \textbf{\textcolor{FemalePurple}{-e}}, \textbf{\textcolor{FemalePurple}{-ster}}) used to refer exclusively to female referents. The \textbf{subject of the sentence} (e.g.\ \textit{man} `man’, \textit{vrouw} `woman’) provides the explicit contextual gender and is referred to as the \textbf{target word}. The target words are always gender-referential (see Fig.~\ref{fig:example_sentences_simple}).
Throughout the article, we use colour conventions in figures to visually distinguish forms: \textbf{\textcolor{FemalePurple}{purple}} denotes female-refenential forms, \textbf{\textcolor{MaleOrange}{orange}} denotes male-referential forms, \textcolor{ProSBlue}{blue} highlights pro-stereotypical pairings, and \textbf{\textcolor{AntiSGreen}{green}} highlights anti-stereotypical pairings.

\textbf{
\begin{figure}[htbp]
\centering
\includegraphics[width=0.65\textwidth]{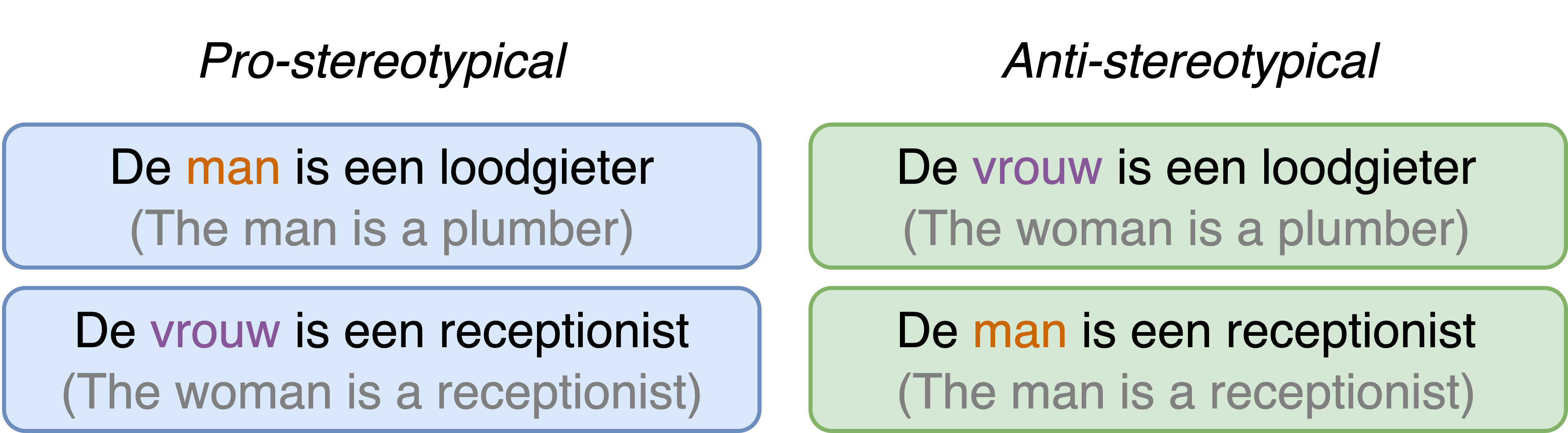}
\caption{Example Dutch sentence pairs used in our projection-based experiments. 
Each sentence follows the template “[TARGET] is een [ATTRIBUTE]” and varies by 
the gender of the subject (man/vrouw) and the stereotype associated with the profession. 
Sentences on the left are \textbf{\textcolor{ProSBlue}{pro-stereotypical}}; those on the right are 
\textbf{\textcolor{AntiSGreen}{anti-stereotypical}}. For clarity, all examples use neutral 
(historically male) profession forms. Explicitly female variants (e.g.\ 
\textit{receptionist\textbf{\textcolor{FemalePurple}{e}}}) are introduced in Section~\ref{sec:results}.}
\label{fig:example_sentences_simple}
\end{figure}
}

To disentangle the effects of grammatical gender and societal stereotypes, we systematically vary three factors: (i) the gender of the target (male/female), (ii) the grammatical gender marking of the attribute (neutral vs.\ female-suffix), and (iii) the stereotype alignment of the target-attribute pair (pro-stereotypical vs. anti-stereotypical). Ungrammatical combinations (e.g., male targets with female-suffix attributes such as \textit{De man is een verpleegster} (`The man is a female nurse')) are excluded. 

Attribute words are drawn from the 2024 Dutch occupational participation dataset \cite{cbs2024beroepen}. Following \cite{bartl-etal-2020-unmasking}, we label each profession as stereotypically male or female based on gender participation.\footnote{\label{fn:cbs}If more than 50\% of the people in the profession are women, we label it as a stereotypically female profession and viceversa.} Generating all possible target–attribute combinations yields 5{,}148 sentences (see Table~\ref{tab:attributes-both-forms}).

\begin{figure}[htbp]
\centering
\textbf{\small Example sentences in the sentence template.} \\[1ex]
\includegraphics[width=0.7\textwidth]{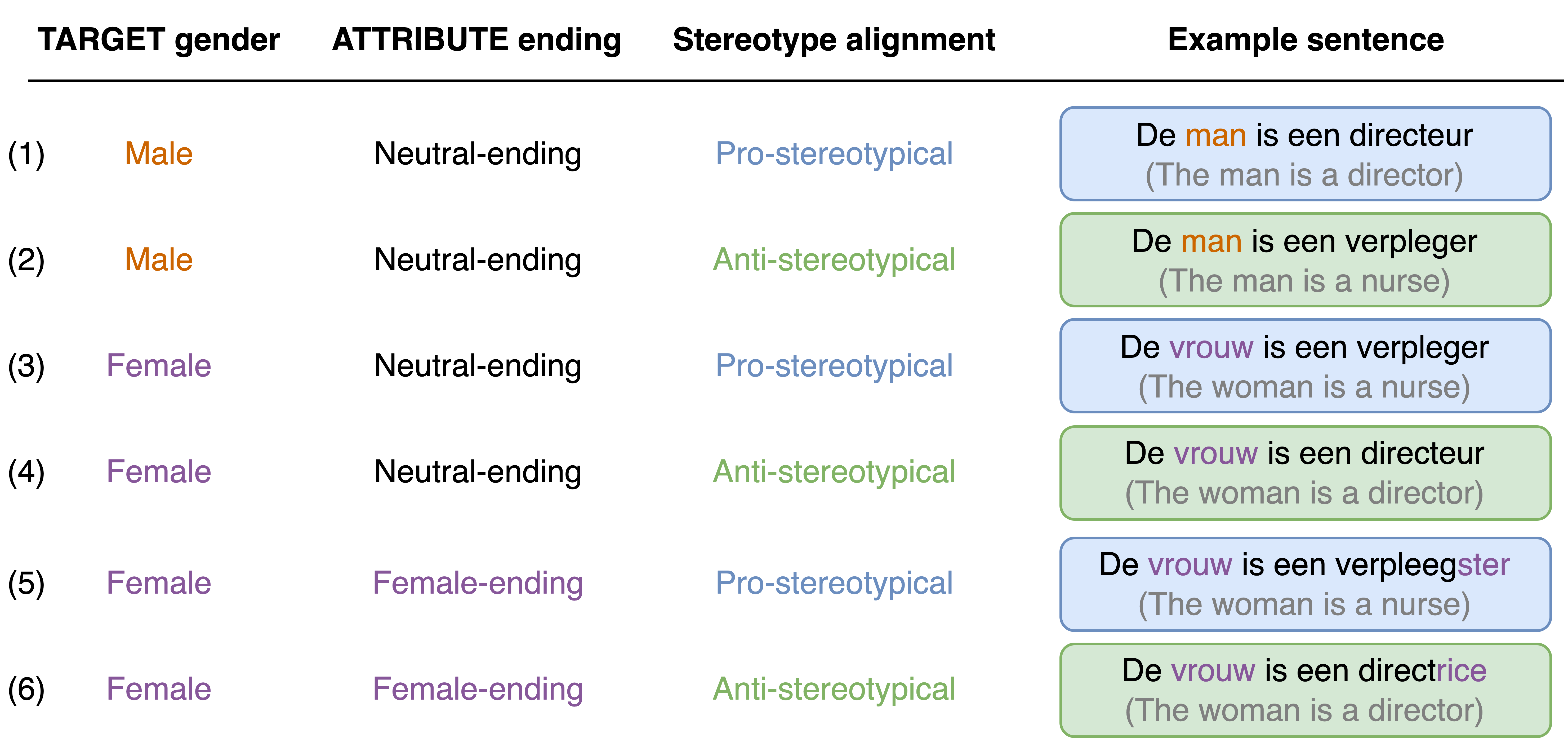}
\caption{Six example Dutch sentences using the template "[TARGET] is een [ATTRIBUTE]" used to test gender bias in our BERT model. The examples systematically vary target gender (\textcolor{MaleOrange}{male} or \textcolor{FemalePurple}{female}), the grammatical gender marking of the attribute (neutral vs. \textcolor{FemalePurple}{female-suffix}), and stereotype alignment (\textcolor{ProSBlue}{pro-stereotypical} vs. \textcolor{AntiSGreen}{anti-stereotypical}). This controlled setup enables isolation of grammatical form, target gender, and societal bias on gender predictions.}
\label{fig:example_sentences}
\end{figure} 


Given a target \( t \in \{\textit{hij (he)}, \textit{zij (she)}, \ldots\} \) and an attribute \( a \in \{\textit{dokter}, \textit{verpleegster}, \ldots\} \)
(dokter = doctor, verpleegster = nurse), we form the sentence \( s = \text{`}t \text{ is een (is a) } a \text{'} \) and extract the contextual embedding of the attribute \( \vec{a} \in \mathbb{R}^{768} \) from the final layer of BERT. The degree to which \( a \) is associated with male or female gender is then estimated by projecting it onto the SVM-derived gender direction \( \vec{g} \) (Section~\ref{sec:gender_subspace}):

\begin{equation}
\text{bias}_{\text{SVM}}(\vec{a}) = \langle \vec{g}, \vec{a} \rangle
\label{eq: gender_projection_sentence}
\end{equation}
Since positive values indicate alignment with the male gender and negative values with the female, we treat the sign as a gender prediction \( \hat{y} \) and compare it to the gender of the target word \( y \). Accuracy is then computed as the proportion of sentences for which \(\hat{y} = y\). Specifically, we evaluate performance across three structured conditions, corresponding to the columns of Figure~\ref{fig:example_sentences}.
\\ \\
\textbf{Phase 1: Stereotype alignment} 
We first compare the SVM’s predictions for pro-stereotypical versus anti-stereotypical sentence constructions. Stereotypical sentences are those in which the target's gender aligns with societal expectations for the attribute, based on labor force data from the CBS\footnotemark[\getrefnumber{fn:cbs}]. Anti-stereotypical sentences deliberately invert these expectations.
\\ \\
\textbf{Phase 2: Target gender} 
Second, we evaluate the SVM's performance as a function of the target's gender. This enables us to investigate whether there is an asymmetry in how the model handles sentences with \textcolor{MaleOrange}{male} versus \textcolor{FemalePurple}{female} subjects, and whether one gender cue exerts less influence on the resulting representations in the presence of stereotypical pressures.
\\ \\
\textbf{Phase 3: Attribute morphology}
Finally, we expand our analysis by incorporating the morphological suffix of each profession. In Dutch, many professions exist in the neutral (or male) form and in the female form (e.g., \textit{verpleger (nurse)} vs. \textit{verpleeg\textbf{\textcolor{FemalePurple}{ster}} (nurse)}). Including both forms allows us to test the influence of the female-marked suffixes in attributes on prediction accuracy.
\\ \\
This allows us to isolate and interpret the different factors contributing to gender bias in contextualized representations. The results for this analysis can be found in Section~\ref{sec:bias}.

\section{Results} \label{sec:results}
The following section presents our findings and addresses the guiding questions of this study: (i) when gender becomes linearly separable during training, (ii) how gender information is distributed across embedding dimensions, (iii) whether stereotypical gender-profession associations dominate the gender information encoded in the embeddings, and (iv) how morphological gender marking affects the extent to which contextual information is encoded.

Before turning to these analyses, we verify that our BERT model converged during training. As reported in Appendix B (p.~\pageref{app:b}), the training and validation loss curves decrease steadily and stabilize over the final epochs. This indicates that BERT has reached a stable internal representation of Dutch, making it suitable for further probing and ensuring that our analysis is not confounded by training artifacts, such as underfitting or overfitting.


\subsection{Gender Bias During Training}
\label{sec:results_emergence}
To answer question (i) (when gender becomes linearly separable during training), we track the classification accuracy of our SVM classifier -- trained on all embedding dimensions -- at each training checkpoint. As shown in Figure~\ref{fig:svm_accuracy_over_time}, accuracy starts around chance level (0.50) and from here  rises steadily throughout the early epochs, surpassing 0.68 by the end of epoch 2. Beyond epoch 20, performance plateaus just above 0.92. This shows that gender information becomes stable and highly separable. While this is  consistent with findings by van der Wal et al. (2022), our results differ in how quick such a high accuracy is reached. Their LSTM already reaches the maximum accuracy at epoch 3, while the accuracy of our SVMs, based on our BERT-based model increases past that point, indicating a more gradual encoding process.

\begin{figure}[htbp]
\centering
\textbf{\scriptsize SVM Accuracy Across BERT Epochs} \\[1ex]
\includegraphics[width=\textwidth]{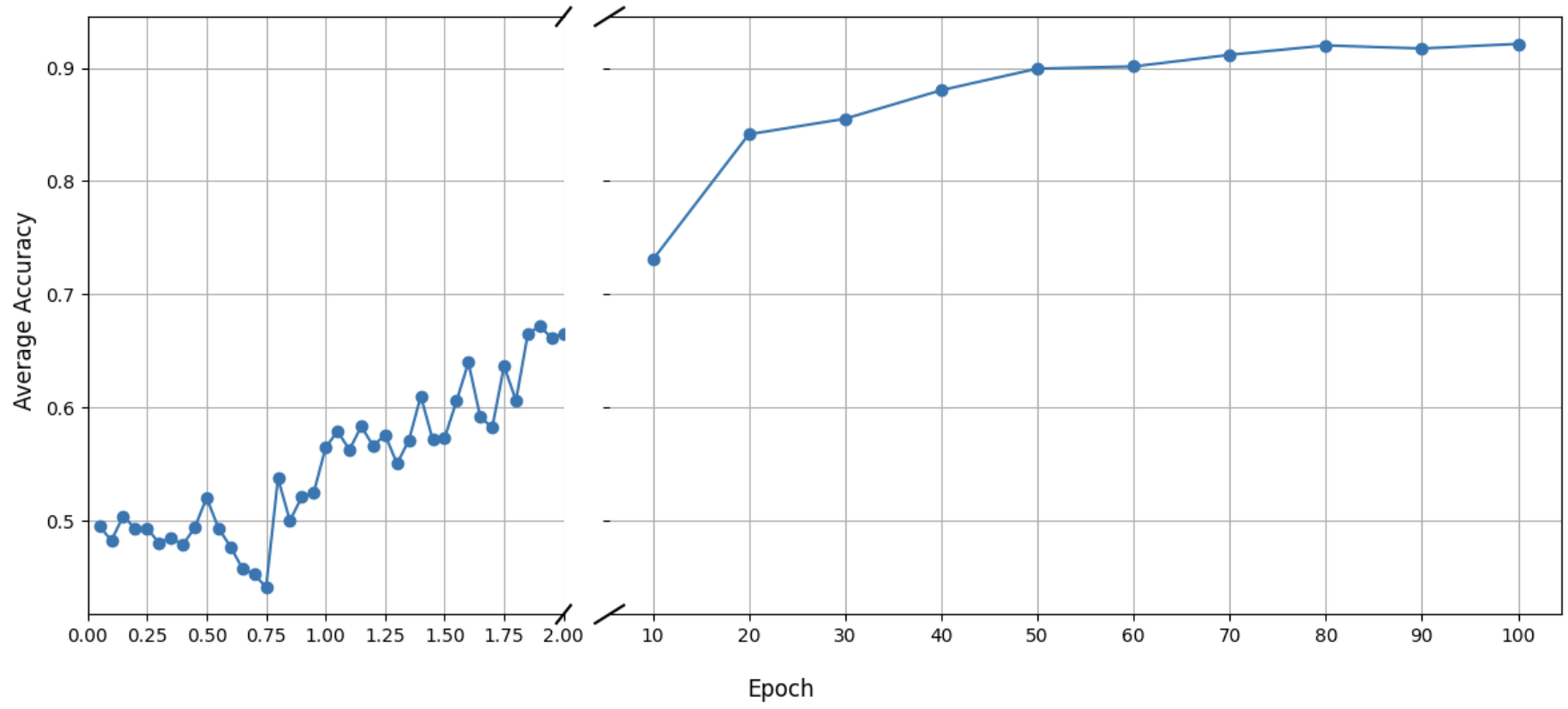}
\caption{SVM classification accuracy over BERT training epochs, using all embedding dimensions. At each BERT checkpoint, a new SVM is fully trained on contextual embeddings to classify gender. Accuracy increases steadily and stabilizes above 0.92. It is only slightly increasing after epoch 20, indicating that gender becomes linearly separable in BERT’s embedding space at that point. \textit{Note: Colors do not reflect the color scheme defined in section \ref{sec:methodology}.}}
\label{fig:svm_accuracy_over_time}
\end{figure} 

We then examine how gender information is distributed across the embedding space, to address question (ii) (how gender information is distributed across embedding dimensions). Figure~\ref{fig:svm_accuracy_over_time_three} extends the previous analysis by comparing three classifiers: an SVM trained on all embedding dimensions, an SVM trained only on the single best-performing dimension (211), and an SVM trained on all dimensions except 211. While the classifier using only dimension 211 reaches an accuracy above 77\% by epoch 100, removing this dimension (``all but best'') barely affects the model's performance, which remains high (above 92\%). This strongly indicates that BERT does not rely on a single ``gender unit''. Instead, the model distributes gender information across more than one dimension.

\begin{figure}[htbp]
\centering
\textbf{\scriptsize SVM Accuracy Across BERT Epochs} \\[0.3ex]
\scriptsize Including the best dimension and all but the best dimensions \\[1ex]
\includegraphics[width=\textwidth]{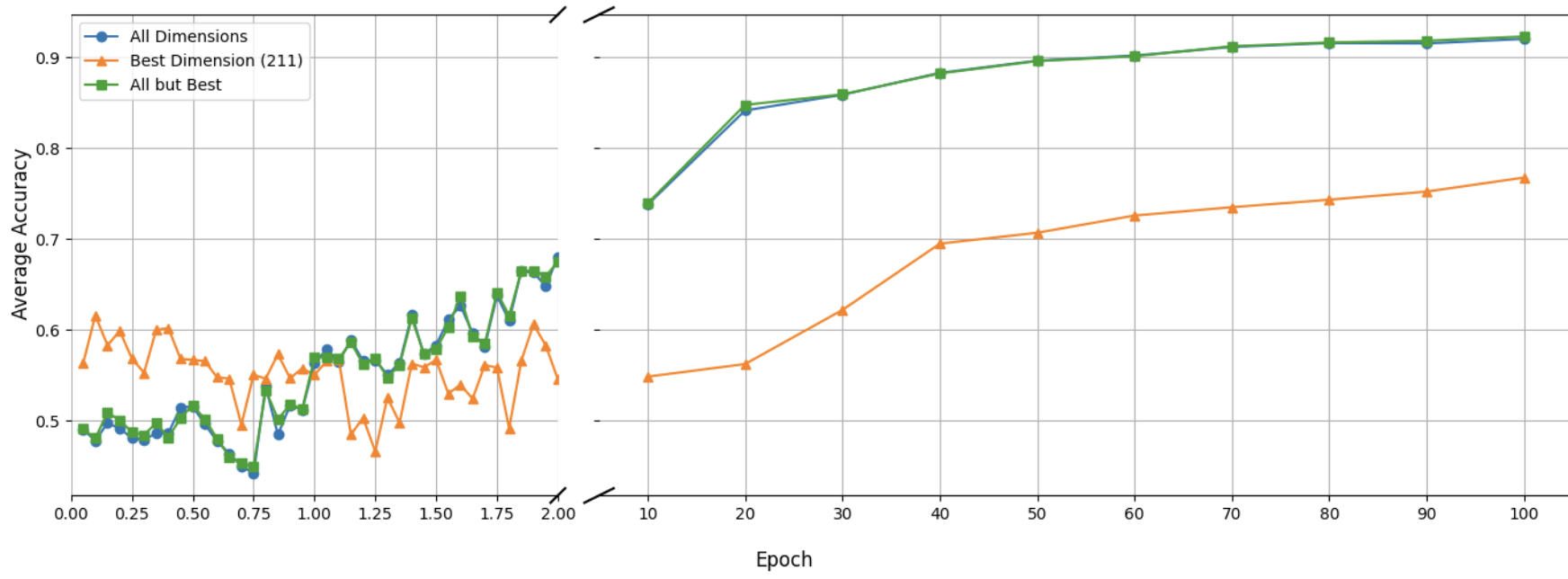}
\caption{SVM classification accuracy over BERT training epochs using all embedding dimensions (blue), only the best-performing dimension (orange, dimension 211), and all dimensions except the best one (green). The high accuracy of the green line shows that removing the top dimension has minimal effect, indicating that gender information is distributed across multiple embedding dimensions. \textit{Note: Colors do not reflect the color scheme defined in section \ref{sec:methodology}.}}
\label{fig:svm_accuracy_over_time_three}
\end{figure} 

Figure~\ref{fig:svm_accuracy_over_time_five} reinforces this by showing the accuracy of classifiers trained on the five most informative individual dimensions. Each of these dimensions (211, 270, 586, 721, and 604) achieves relatively high accuracy on its own, further demonstrating that gender information is distributed across multiple dimensions. This finding aligns with \citeasnoun{ravfogel-etal-2020-null}, who argue that gender is not captured by a single direction but is distributed across many dimensions. Our results extend this claim to contextual embeddings. Similarly, our results also align with and extend \citeasnoun{liang-etal-2020-monolingual}. Although their analysis concerns distribution across layers and heads rather than embedding dimensions and is derived from debiasing effectiveness rather than subspace construction, they too provide evidence that gender encoding in BERT is distributed rather than localized.

\begin{figure}[htbp]
\centering
\textbf{\scriptsize SVM Accuracy Across BERT Epochs} \\[0.3ex]
\scriptsize Including the 5 best performing dimensions \\[1ex]
\includegraphics[width=\textwidth]{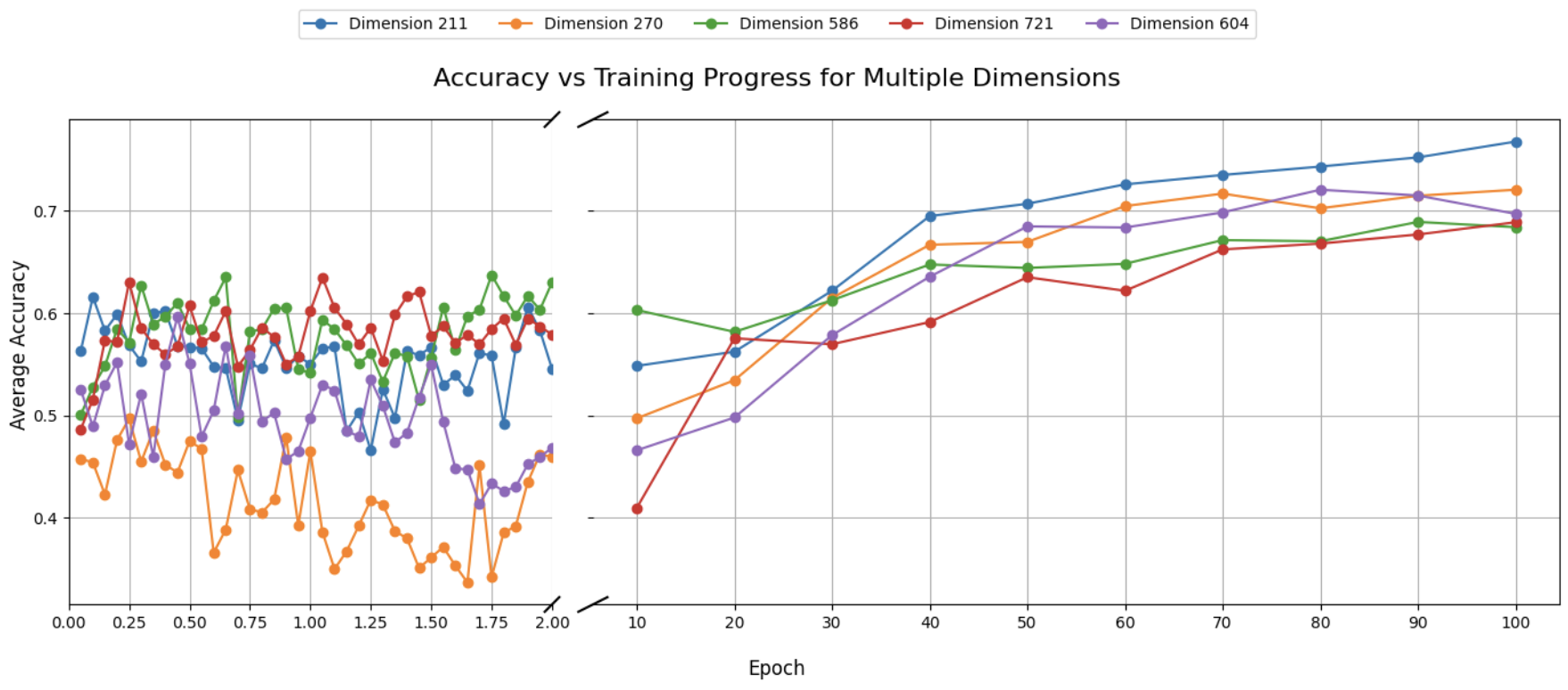}
\caption{SVM classification accuracy over BERT training epochs using the five best-performing individual embedding dimensions. While each dimension achieves moderate accuracy on its own, none dominates entirely, confirming that gender information is distributed across multiple dimensions rather than concentrated in a single ``gender unit''.}
\label{fig:svm_accuracy_over_time_five}
\end{figure} 



Finally, heatmaps of cluster-wise mean recall values for each gender shown in Figures~\ref{fig:feminine-recall} and~\ref{fig:masculine-recall} reveal a pronounced asymmetry in how gender is encoded. While overall linear separability of gender rises early in training (as shown in Figure~\ref{fig:svm_accuracy_over_time}), the recall distributions indicate that this separability benefits the \textit{Male} class more than the \textit{Female} class. Specifically, the \textit{Male} class recall values are high across many dimensions. The recall values for the \textit{Female} class, however, show wider dispersion, with several dimensions yielding very low recall ($<$0.1). This suggests that information associated with the \textit{Male} class is more uniformly distributed across the embedding space, whereas information associated with the \textit{Female} class is concentrated in fewer, more specific dimensions (the highest dimension having a recall of 0.74). These patterns suggests that the representation associated with the \textit{Male} and \textit{Female} classes are not encoded as two opposite poles of a single semantic dimension. Rather, the \textit{Male} class appears to function as a more diffuse, unmarked default, while the \textit{Female} one is encoded in a more specific and concentrated way, consistent with broader notions of markedness and default encoding which is in line with similar asymmetries observed in prior work on gender representations in neural language models \cite{vanderwal2022birthbiascasestudy,manna2025payingattentionherinvestigating}.


\begin{figure}[htbp]
    \centering
    \includegraphics[width=\textwidth]{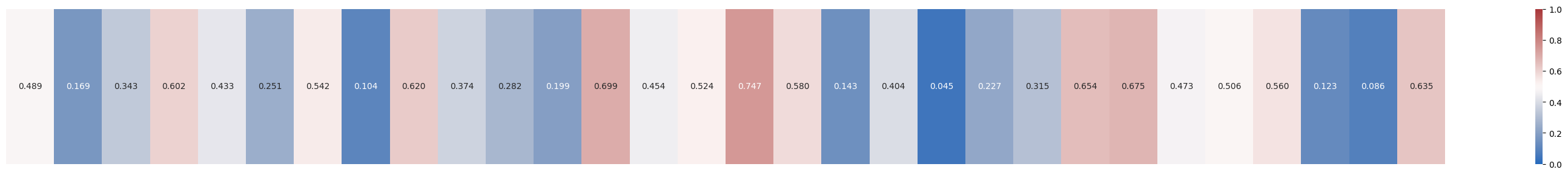}
    \caption{Clustered average recall for the \textit{Female} class across embedding dimensions at BERT epoch 50, with dimensions grouped into 30 K-Means clusters. High-recall clusters are sparse and localized, indicating that female gender information is concentrated in a limited number of dimensions.}
    \label{fig:feminine-recall}
\end{figure} 
\begin{figure}[htbp]
    \centering
    \includegraphics[width=\textwidth]{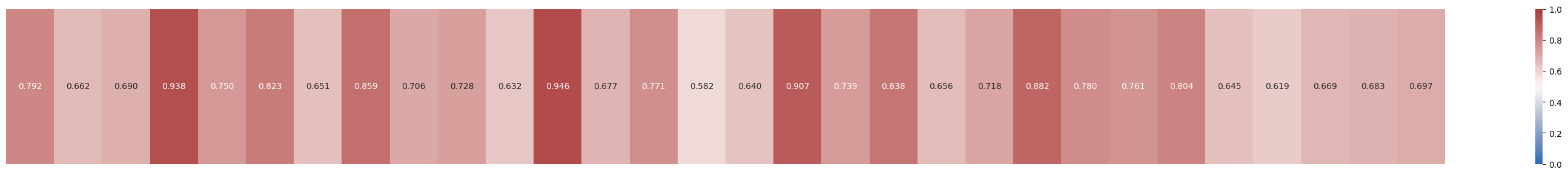}
    \caption{Clustered average recall for the \textit{Male} class across embedding dimensions at BERT epoch 50, with dimensions grouped into 30 K-Means clusters. Recall is higher and more evenly distributed, suggesting that male gender information is encoded more diffusely across the embedding space.}
    \label{fig:masculine-recall}
\end{figure} 
}

\subsection{Bias in sentence contexts} \label{sec:bias}
\begin{figure}[htbp]
\centering
\textbf{\scriptsize Prediction accuracy by stereotype alignment.} \\[0.3ex]
\includegraphics[width=0.65\textwidth]{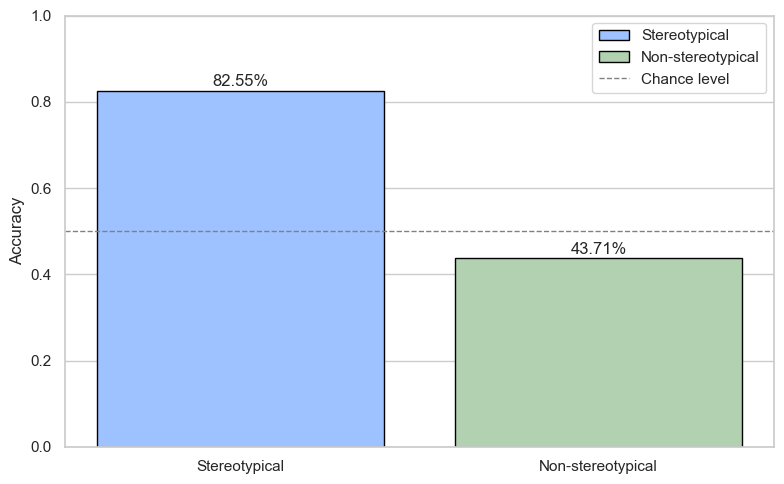}
\caption{Accuracy difference between pro-stereotypical and anti-stereotypical gender–profession pairs. Accuracy is substantially higher for stereotypical sentences (82.55\%) than for anti-stereotypical ones (43.71\%), indicating a strong alignment with societal gender stereotypes. The dotted line marks chance level (50\%).}
\label{fig:stereotype_accuracy_2}
\end{figure}

\begin{figure}[htbp]
\centering
\textbf{\scriptsize Prediction accuracy by stereotype alignment and gender} \\[0.6ex]
\includegraphics[width=0.65\textwidth]{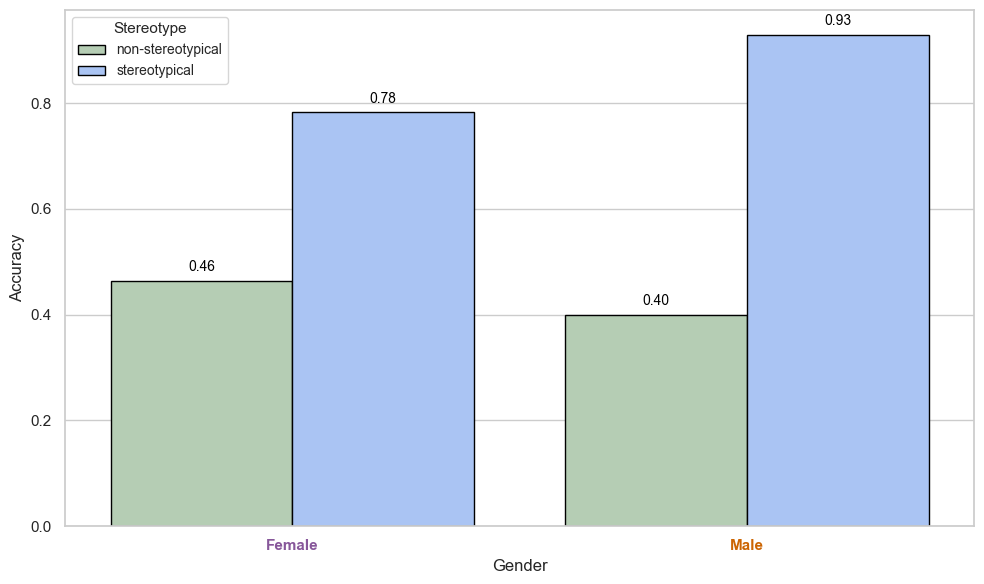}
\caption{Gender prediction accuracy split by target gender and stereotype alignment. Accuracy is higher for stereotypical cases in both genders, but the effect is more pronounced for male targets (93\% vs. 40\%) than for female targets (78\% vs. 46\%), indicating a stronger bias toward male-stereotypical associations.}
\label{fig:gender_stereotype_split}
\end{figure}

\begin{figure}[htbp]
\centering
\textbf{\scriptsize Prediction accuracy by stereotype alignment, gender and lexical form} \\[0.3ex]
\includegraphics[width=0.65\textwidth]{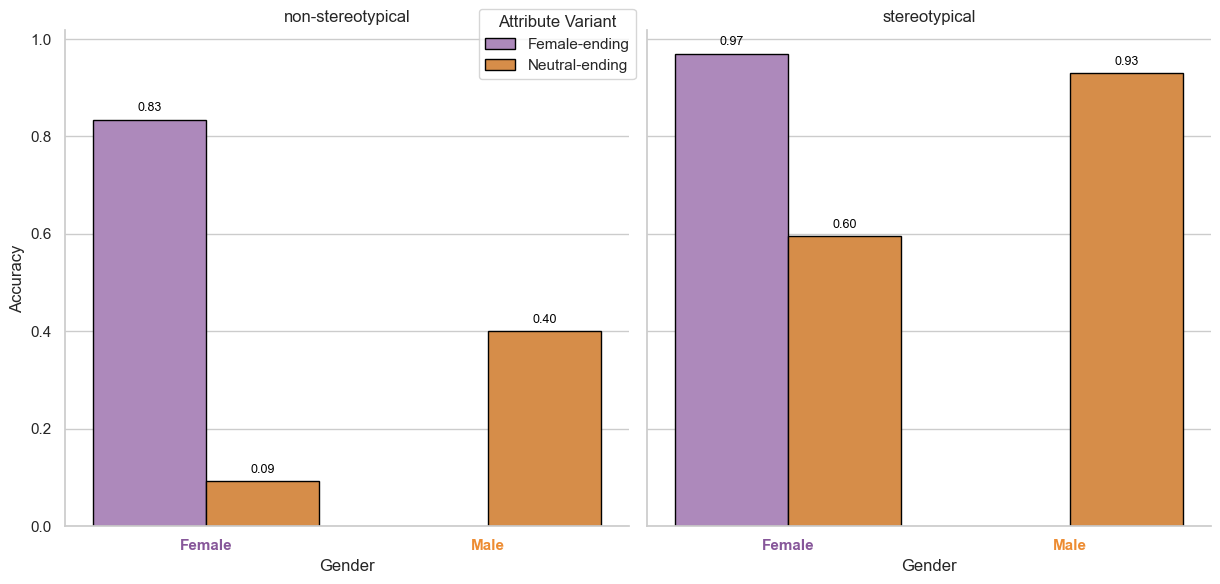}
\caption{Prediction accuracy across profession variants, split by target gender, stereotype alignment, and female-marked attributes. For female targets, female-marked attribute suffixes dramatically increase accuracy, especially in anti-stereotypical contexts (83\% vs. 9\%).}
\label{fig:variant_combination}
\end{figure}

To answer questions (iii) (whether stereotypical gender-profession associations dominate the gender information encoded in the embeddings) and (iv) (how morphological gender marking affects the extent to which contextual information is encoded.), we analyse how gendered targets and profession attributes interact in the embedding space. To this end, Table~\ref{tab:accuracy} summarises the average accuracy of our SVM classifiers across three structured conditions: stereotype alignment, target gender, and attribute morphology. We distinguish between the referential gender of the target (e.g., vrouw vs. man) and the morphological marking of the attribute (e.g., verpleegster vs. verpleger). \\ \\
\textbf{Stereotype alignment} As illustrated in Figure~\ref{fig:stereotype_accuracy_2}, overall accuracy is substantially higher for stereotypical pairings (82.5\%) than for anti-stereotypical ones (43.7\%; Table~\ref{tab:accuracy}). This discrepancy indicates a strong distributional bias, suggesting that the resulting representations are primarily shaped by learned statistical associations rather than available contextual (gender) cues. \\ \\
\textbf{Target gender}
Further breaking the results down by target gender reveals that this effect is markedly stronger for male targets (Figure~\ref{fig:gender_stereotype_split}). While accuracy for female targets drops by 32 points in anti-stereotypical settings, male targets see a 53-point decrease. This suggests that antistereotypical sentences involving male targets disrupt the model’s internal gender logic more strongly, reinforcing the finding that the model prioritizes pre-trained gender associations over contextual information.\\ \\
\textbf{Attribute morphology} 
We observe that the moderately high accuracies for female targets in Figure~\ref{fig:gender_stereotype_split} are largely driven by female marked profession forms. As shown in Figure~\ref{fig:variant_combination}, female suffixes provide a strong gender signal to the classifier: in anti-stereotypical contexts these forms yield 83\% accuracy, whereas neutral attributes drop sharply to only 9\%. A similar pattern appears in stereotypical contexts, where female suffix forms achieve 97\% accuracy compared to 60\% for neutral forms.

These differences show that the aggregate accuracies for female-target sentences in Figure \ref{fig:gender_stereotype_split} (46\% and 78\%) mask an important effect: our classifier attains high accuracy for female-target sentences only when the underlying sentence contains an explicit morphological marking. When the attribute is morphologically neutral, accuracy drops sharply. This suggests that the model's internal representation of gender is driven largely by this surface-level morphological cue rather than the integration of gendered information from the surrounding sentence.

\begin{table}[htbp]
\centering
\renewcommand{\arraystretch}{1.2}
\begin{tabularx}{\textwidth}{>{\raggedright\arraybackslash}X c c}
\toprule
\textbf{Comparison} & \textbf{Accuracy (Avg)} & \textbf{Difference} \\
\midrule
\textbf{General Stereotype Effect} & & \\
All: Stereotypical vs Non-stereotypical & 0.825 vs 0.437 & \textbf{+0.388*} \\
\midrule
\textbf{Within-Gender Comparisons} & & \\
Female: Stereotypical vs Non-stereotypical & 0.780 vs 0.460 & \textbf{+0.320*} \\
Male: Stereotypical vs Non-stereotypical & 0.930 vs 0.400 & \textbf{+0.530*} \\
\midrule
\textbf{Cross-Gender Comparisons (Same Context)} & & \\
Non-stereotypical: Female vs Male & 0.460 vs 0.400 & \textbf{+0.060} \\
Stereotypical: Female vs Male & 0.780 vs 0.930 & \textbf{-0.150} \\
\midrule
\textbf{Effects of female-marked attributes (Female-suffix vs Neutral)} & & \\
Female Anti-stereotypical: Female suffix vs Neutral & 0.830 vs 0.090 & \textbf{+0.740*} \\
Female Pro-stereotypical: Female suffix vs Neutral & 0.970 vs 0.600 & \textbf{+0.370*} \\
\bottomrule
\end{tabularx}
\caption{Average model accuracy across stereotype alignment, target gender, and female-marked attributes. The “Comparison” column describes the evaluated conditions, “Accuracy (Avg)” reports mean accuracy per condition, and “Difference” reflects accuracy changes between contrasts. Stereotypical gender–profession pairings show higher accuracy than anti-stereotypical ones, particularly for male targets. Female-marked attributes substantially increase accuracy relative to neutral forms, even in anti-stereotypical contexts. Asterisks indicate statistically significant differences ($p < 0.05$, two-proportion $z$-tests; Appendix E, Table \ref{tab:ztests}).}
\label{tab:accuracy}
\end{table}


These findings overall strongly support the idea that, when an attribute with a \textbf{female marked} suffix is used, the resulting embeddings align strongly with the female side of the gender subspace, regardless of pronoun or stereotype.
When a \textbf{neutral} attribute is used, BERT's representation is more ambiguous; in these cases, the model tends to default to the male side of the learned gender subspace
, especially in stereotypical contexts. This observation does not make claims about real-world gender bias within the professions themselves, but rather reflects the way gender is distributed and marked in the training corpus used by the model overall. It directly informs the answer to (iv) (how morphological gender marking affects the extent to which contextual information is encoded), but does not change the conclusion to (iii) (whether stereotypical gender-profession associations dominate the gender information encoded in the embeddings), which still indicates the presence of bias.

\section{Discussion} \label{sec:discussion}

\vspace{1em}

\subsection{Positioning the findings in relation to prior work} \label{sec:findings_compare}

This research contributes to the literature by investigating how gender bias emerges and evolves within a Dutch BERT model trained entirely from scratch. Most previous research primarily focused on English. This study, on the other hand, addresses the complexities of morphologically gender-marked Dutch. By examining contextual embeddings and constructing a contextual gender subspace using linear Support Vector Machines (SVMs) across multiple training checkpoints, this research  shows when and how gender information is encoded during model training. Moreover, by  projecting profession terms within controlled contexts on the gender subspace, this study provides new insights into how explicitly gendered cues within these controlled contexts interact with intrinsic model biases.

Our findings extend prior work on gender bias in contextualised language models, particularly by making explicit how limited the corrective power of clear contextual cues can be. Previous studies using sentence templates have shown that stereotypical sentences receive systematically higher likelihood scores than anti-stereotypical ones \cite{kurita-etal-2019-measuring,nangia-etal-2020-crows}, but these studies primarily evaluate model outputs or likelihoods. By projecting contextualised profession embeddings onto a learned gender subspace, our analysis shows that this asymmetry is already present at the level of internal representations: even when a sentence contains an unambiguous gender cue (e.g., a female subject), the resulting profession embedding often remains aligned with its stereotypical gender.

At the same time, we uncover an asymmetry in how gender information is distributed across the embedding space, with male information being more diffusely encoded, whereas female signals appear to be concentrated in a more limited subset of embedding dimensions. This pattern points to a default-to-male tendency, in which the male gender seems to function as a more general, unmarked default, while female signals appear to be explicitly encoded in localized dimensions. Similar asymmetries have been observed in prior work on the encoding of gender information in language models \cite{vanderwal2022birthbiascasestudy,manna2025payingattentionherinvestigating}. 

A related pattern emerges when considering the role of morphological gender marking. Prior work on morphologically richer languages, such as German, shows that female markings on profession nouns have a strong influence on model predictions \cite{bartl-etal-2020-unmasking}. We observe something comparable in Dutch: female suffixes reliably lead to female interpretations, whereas the generic forms tend to default to male ones. Importantly, however, this does not indicate successful contextual integration. Rather, it shows that models respond strongly to morphological markers, while still failing to integrate explicit cues such as gendered subjects. Overall, prediction accuracy is consistently skewed toward male interpretations, benefiting male-coded representations across both stereotypical and anti-stereotypical contexts. 

Nonetheless, this closely relates to the structural properties of Dutch gender morphology outlined in Section~\ref{sec:Dutch}, where ongoing neutralisation efforts have led to historically male, morphologically unmarked (profession) forms being often preferred over explicitly female ones. The representations learned by the model therefore appear to mirror not only statistical regularities in the training data, but also the morphological asymmetries of the Dutch language itself.


\subsection{Context as constraint rather than corrective}
Contextualised word embeddings are designed to integrate local contextual cues, allowing word meanings to shift depending on context. Our results challenge this view in a more fundamental way. Rather than observing context as a corrective force that overrides prior gender associations, we find that contextual cues often fail to counteract prior gender assumptions. In many cases, once a profession is encoded as male- or female-typed, the local context alone is often insufficient to reverse this, since it competes with the (broader) learned context inherited from the training data.

As discussed in Section~\ref{sec:Dutch}, this limitation is particularly pronounced in Dutch, where references to women are often expressed via explicit suffixation, while the morphologically unmarked profession forms are used both generically and for male referents. This asymmetry in the Dutch language, and therefore in the training data, causes the unmarked form to function as a male default in practice, giving male-coded representations a structural advantage in training and reducing the corrective impact of contextual gender cues even further. As a result, contextual gender cues that conflict with stereotypical associations may exert a weaker influence on the resulting embeddings.

This raises a broader conceptual question for contextual language models: If contextual cues do not consistently override learned gender priors, to what extent can these models be said to dynamically integrate context at all? Rather than updating representations in response to new information in a sentence, our results suggest that the resulting representations are primarily shaped by associations established during training. More broadly, this means that context can only shift gender predictions when similar context have been sufficiently seen during training. But, what does it mean for a model to be context-sensitive if its ability to adapt depends so strongly on whether a given context has been seen often enough during training? Especially since this means it does not integrate new information (enough)?

It is important to note that our analysis probes linear separability in contextual word embeddings learned by our BERT model and does not directly measure model performance on downstream tasks; as such, our conclusions are limited to the representational level captured by the applied probes.


\vspace{0.2cm}

\section{Conclusion} \label{sec:conclusion}

This study showed how gender information develops in a Dutch BERT model trained from scratch, and whether explicit contextual cues can counteract the gender patterns learned from data. By tracking gender encoding across training checkpoints and projecting profession embeddings from controlled sentences onto learned gender subspaces, we analyzed when gender becomes linearly separable, how it is distributed across the embedding space, and how it responds to context. We now return to our central question: \textit{How does a Dutch BERT model encode gender information during training, and to what extent do its final profession representations reflect societal gender bias?}

We find that gender becomes clearly linearly separable relatively early in training, stabilizing around epoch 20. However, this separability is uneven. Male information is spread across many embedding dimensions, while female information is concentrated in a smaller number of dimensions. As a result, male interpretations have a built-in advantage: even when a profession appears in an explicitly female context, its representation often remains male-coded.

Our projection-based analysis further shows that contextual cues alone are often not enough to override these learned associations. Stereotypical gender–profession pairings are classified with higher accuracy by the SVM when projected onto the learned gender subspace than anti-stereotypical ones. Explicitly female suffixes have a strong effect and reliably push representations toward the female side of the gender subspace, even in conflicting contexts. This suggests that the resulting representations are primarily shaped by clear morphological markers and learned frequency patterns, rather than by contextual cues in the sentence.

These results have important implications for the use of Transformer-based models in downstream tasks. If explicit context can be overridden by biased defaults at the representational level, applications such as translation, coreference resolution, or dialogue systems may reproduce gender stereotypes even when the input clearly contradicts them. More generally, our study suggests that claims about context sensitivity in language models should be treated with caution, especially for grammatically gendered languages like Dutch.

\section*{Limitations} \label{sec:limitations_future_research}

While our results clearly show that profession terms with female-marked suffixes lead to significantly higher gender prediction accuracy, even in anti-stereotypical contexts, we cannot claim with certainty that these effects are caused solely by the morphological form. We can claim correlation but not causation.

This is because our gender subspace reflects both morphological features and statistical co-occurrence patterns from the corpus. Therefore, morphological gender markers and societal stereotype frequency are entangled in the model’s embeddings. This study does not fully separate these components within the embedding space. Our method relies on a single learned gender direction, which may blend morphological cues (the female-marked suffixes) with semantic gender associations (the stereotypical expectations about professions). \citeasnoun{zhou-etal-2019-examining} address this by constructing two orthogonal subspaces: one encoding grammatical gender and the other semantic gender. This separation enables more precise analysis of what kind of gender signal is present in each embedding. However, their approach is not directly transferable to Dutch. Unlike languages such as Spanish or French, Dutch does not apply systematic gender marking to inanimate nouns. \citeasnoun{zhou-etal-2019-examining} specifically use these inanimate nouns to create the grammatical gender subspace, since these are different only in their gendered marking and not structurally in other semantics. Dutch only has this distinction in profession terms and not in inanimate nouns, making it difficult to construct a grammatical gender subspace. To address this, another strategy would be to examine how gender information is encoded across different layers of the model. It is well established that lower layers in transformer architectures tend to capture more syntactic and surface-level features, while higher layers encode increasingly semantic and contextual information \cite{jawahar-etal-2019-bert,tenney-etal-2019-bert}. In our setup, we extract embeddings exclusively from the final hidden layer, which likely mixes morphological gender signals (e.g., the presence of a female suffix) with semantic gender associations (e.g., the stereotypical gender of a profession). A layer-wise analysis could help tease these apart. Such a decomposition would offer an alternative, or complement, to the dual-subspace approach of \citeasnoun{zhou-etal-2019-examining}, and would be particularly valuable for a language like Dutch, where morphological and semantic gender are difficult to isolate.

We suggest two directions future research should focus on. First, thus a layer-wise analysis of gender encoding, which could help disentangle morphological from semantic gender signals. Second, future work could focus on developing a synthetic grammatical gender axis for Dutch by creating controlled nonce words with morphological gender markers. This is inspired by \citeasnoun{arps-etal-2024-multilingual}, who created SPUD, a framework for producing syntactically sound but semantically nonsensical sentences through controlled lexical substitution. The development of a synthetic grammatical gender axis for Dutch that is free of stereotyped associations may be made possible by adapting such an approach to produce pseudo-professions with gendered morphological suffixes (for example, \textit{blorateur} vs. \textit{bloratrice}). These terms are guaranteed to have no stereotypical associations, but do differ in their morphological suffix. This enables creating a grammatical gender subspace for Dutch, despite Dutch not having gendered inanimate nouns like French or Spanish.

\newpage

\section*{Appendix A: Architectual overview of the BERT model} \label{app:a}

\section*{Appendix B: Bert training loss} \label{app:b}

\begin{figure}[htbp]
\centering
\textbf{\scriptsize Training and Validation Loss over Epochs} \\[1ex]
\includegraphics[width=0.57\textwidth]{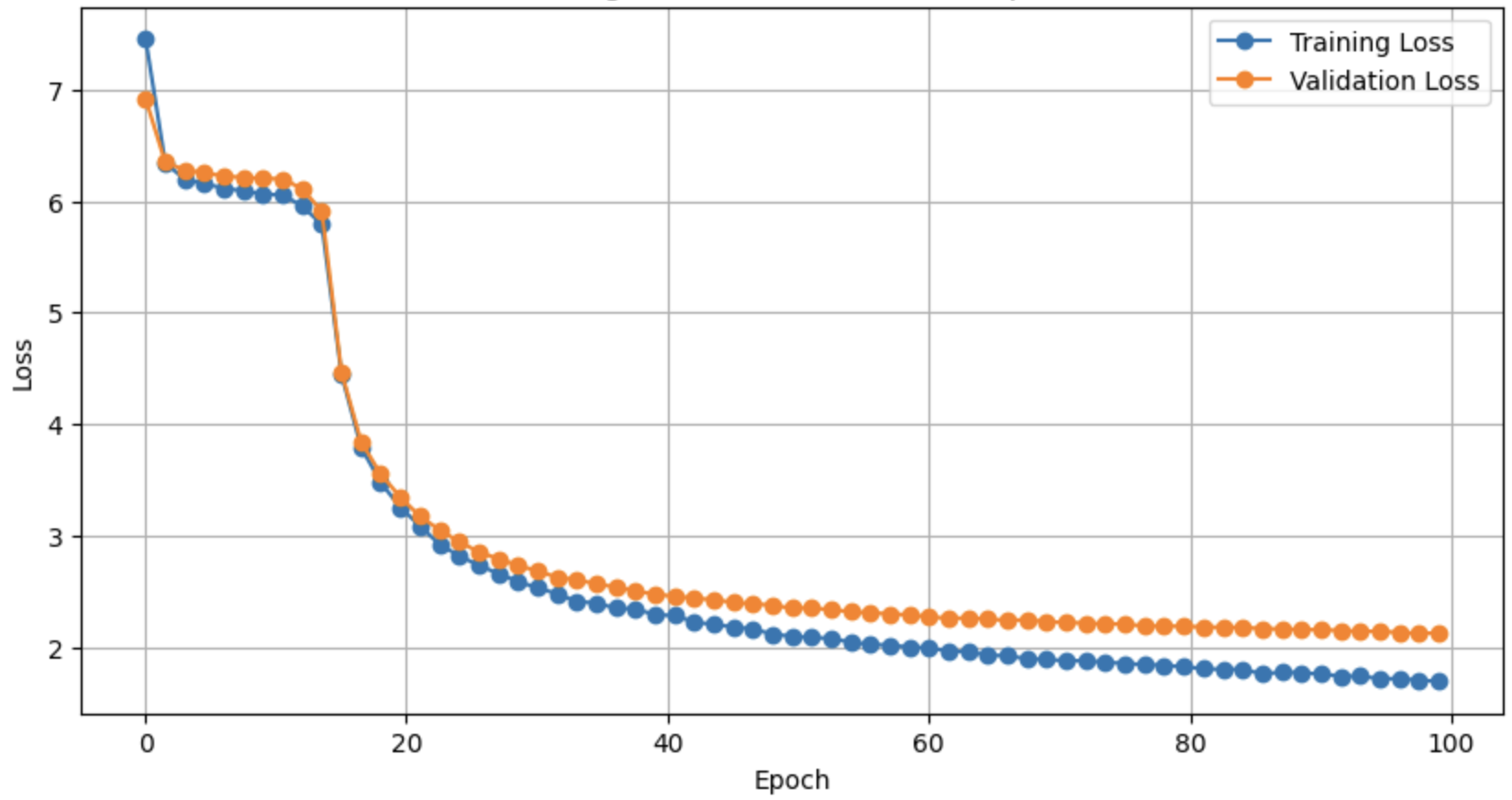}
\caption{Training and validation loss curves across 100 epochs for the Dutch BERT model with the MLM objective. There is a steep drop around epoch 20. The alignment between training and validation loss suggests the model is not overfitting. \textit{Note: Colors do not reflect the color scheme defined in section \ref{sec:methodology}.}}
\label{fig:Bert_loss}
\end{figure}

Figure~\ref{fig:Bert_loss} shows the training and validation loss across 100 epochs of masked language modeling (MLM). Both curves show a steep decline early on. The validation loss drops rapidly in the first 20 epochs, with a sudden drop between epoch 16 and 20. After this, loss slowly decreases. By the final epoch, the model achieves a validation loss of 2.23, indicating that it has converged reasonably well.
This suggests the model has developed a stable internal representation of Dutch, making it suitable for further probing and good to use for our research.

\section*{Appendix C: Target terms} \label{app:c}

\begin{table}[H]
\centering
\footnotesize
\renewcommand{\arraystretch}{0.95}
\setlength{\tabcolsep}{5pt}
\begin{tabular}{lll}
\toprule
\textbf{English (van der Wal et al.)} & \textbf{Dutch (this paper)} & \textbf{Not Included} \\
\midrule
man – woman         & man – vrouw             & \\
boy – girl          & jongen – meisje         & \\
guy – gal           & gozer – meid            & \\
gentleman – lady    & gentleman – dame        & \\
lord – lady         & lord – lady             & \\
Mister – Miss       & mister – miss           & \\
Mr. – Ms./Mrs.           & mr – ms                 & \\
male – female       & mannelijk – vrouwelijk  & \\
masculine – feminine& mannelijk – vrouwelijk  & \\
\midrule
king – queen        & koning – koningin       & \\
prince – princess   & prins – prinses         & \\
Baron – Baroness    & baron – barones         & \\
duke – duchess      & hertog – hertogin       & \\
monk – nun          & monnik – non            & \\
wizard – witch      & ---                     & omitted \\
landlord – landlady & ---                     & omitted \\
\midrule
he – she            & ---                     & removed (zij = she/they) \\
him – her           & ---                     & removed (haar = her/hair) \\
himself – herself   & hijzelf – zijzelf       & \\
his – her/hers          & ---                     & removed (haar = her/hair) \\
\midrule
father – mother         & vader – moeder         & \\
dad – mum               & papa – mama            & \\
brother – sister        & broer – zus            & \\
nephew – niece          & neef – nicht           & \\
uncle – aunt            & oom – tante            & \\
grandfather – grandmother & opa – oma           & \\
son – daughter          & zoon – dochter         & \\
grandson – granddaughter & kleinzoon – kleindochter & \\
son-in-law – daughter-in-law & schoonzoon – schoondochter & \\
stepfather – stepmother & stiefvader – stiefmoeder & \\
stepson – stepdaughter  & stiefzoon – stiefdochter & \\
father-in-law – mother-in-law & schoonvader – schoonmoeder & \\
bridegroom/groom – bride      & bruidegom – bruid      & \\
husband – wife          & man – vrouw            & \\
godfather – godmother   & peetvader – peetmoeder & \\
\bottomrule
\end{tabular}
\caption{Comparison of English gendered word pairs used by 
\protect\citeasnoun{vanderwal2022birthbiascasestudy} and the Dutch
equivalents used in this paper. Omissions are due to polysemy
(e.g., \textit{zij}, \textit{haar}) or limited corpus coverage.}
\label{tab:gender_pairs}
\end{table}

\begin{table}[htbp]
\centering
\begin{tabular}{ll|ll}
\textbf{Male Word (EN)} & \textbf{Count} & \textbf{Female Word (EN)} & \textbf{Count} \\
\hline
man (man) & 200 & vrouw (woman) & 200 \\
broer (brother) & 200 & zus (sister) & 200 \\
zoon (son) & 200 & dochter (daughter) & 200 \\
neef (nephew/cousin) & 200 & nicht (niece/cousin) & 200 \\
vader (father) & 200 & moeder (mother) & 200 \\
opa (grandfather) & 200 & oma (grandmother) & 200 \\
kleinzoon (grandson) & 200 & kleindochter (granddaughter) & 200 \\
grootvader (grandfather) & 200 & grootmoeder (grandmother) & 200 \\
oom (uncle) & 200 & tante (aunt) & 200 \\
papa (dad) & 200 & mama (mom) & 200 \\
jongen (boy) & 200 & meid (girl) & 200 \\
jongetje (little boy) & 200 & meisje (little girl) & 200 \\
schoonvader (father-in-law) & 174 & schoonmoeder (mother-in-law) & 200 \\
schoonzoon (son-in-law) & 149 & schoondochter (daughter-in-law) & 129 \\
stiefvader (stepfather) & 61 & stiefmoeder (stepmother) & 44 \\
stiefzoon (stepson) & 10 & stiefdochter (stepdaughter) & 6 \\
peetvader (godfather) & 14 & peetmoeder (godmother) & 3 \\
bruidegom (groom) & 189 & bruid (bride) & 200 \\
meneer (mister/sir) & 200 & mevrouw (miss/madam) & 200 \\
mijnheer (gentleman/sir) & 200 & dame (lady) & 200 \\
mister (mister) & 200 & miss (miss) & 150 \\
mr (Mr) & 200 & ms (Ms) & 6 \\
prins (prince) & 200 & prinses (princess) & 200 \\
koning (king) & 200 & koningin (queen) & 200 \\
lord (lord) & 71 & lady (lady) & 80 \\
baron (baron) & 136 & barones (baroness) & 8 \\
hertog (duke) & 176 & hertogin (duchess) & 37 \\
monnik (monk) & 120 & non (nun) & 200 \\
hijzelf (himself) & 200 & zijzelf (herself) & 167 \\
mannelijk (male) & 200 & vrouwelijk (female) & 200 \\
gozer (dude/bloke) & 70 & wijf (woman, derogatory) & 168 \\
\hline
\textbf{Total} & \textbf{5170} & & \textbf{4798} \\
\end{tabular}
\caption{List of gendered word pairs (male vs. female) used to extract contextual embeddings from the BERT model. These embeddings form the basis for training the SVM to construct the gender subspace. The table includes the Dutch words, their English translations, and the number of contextual instances sampled from the corpus for each term.}
\label{tab:embedding_pairs}
\end{table}

\begin{table}[htbp]
\centering
\begin{tabular}{ll|ll}
\textbf{Male Word (EN)} &  & \textbf{Female Word (EN)} &  \\
\hline
Hij (he) & & Zij (she) & \\
De man (the man) & & De vrouw (the woman) & \\
De broer (the brother) & & De zus (the sister) & \\
De zoon (the son) & & De dochter (the daughter) & \\
De neef (the nephew/cousin) & & De nicht (the niece/cousin) & \\
De vader (the father) & & De moeder (the mother) & \\
De opa (the grandfather) & & De oma (the grandmother) & \\
De kleinzoon (the grandson) & & De kleindochter (the granddaughter) & \\
De grootvader (the grandfather) & & De grootmoeder (the grandmother) & \\
De oom (the uncle) & & De tante (the aunt) & \\
De papa (the dad) & & De mama (the mom) & \\
De jongen (the boy) & & De meid (the girl) & \\
Het jongetje (the little boy) & & Het meisje (the little girl) & \\
De schoonvader (the father-in-law) & & De schoonmoeder (the mother-in-law) & \\
De schoonzoon (the son-in-law) & & De schoondochter (the daughter-in-law) & \\
De stiefvader (the stepfather) & & De stiefmoeder (the stepmother) & \\
De stiefzoon (the stepson) & & De stiefdochter (the stepdaughter) & \\
De peetvader (the godfather) & & De peetmoeder (the godmother) & \\
De bruidegom (the groom) & & De bruid (the bride) & \\
De meneer (Sir) & & De mevrouw (Madam) & \\
Mijnheer/De heer (gentleman) & & De dame (the lady) & \\
mister (mister) & & miss (miss) & \\
Mr (Mr) & & Ms (Ms) \\
Mister (Mister) & & Miss (Miss) & \\
Hijzelf (himself) & & Zijzelf (herself) & \\
De gozer (the dude/bloke) & &  Het wijf (the woman — derogatory) & \\
De kerel (the guy/bloke) & & -- & \\
\end{tabular}
\caption{
Target words used in the sentence template “[TARGET] is een [ATTRIBUTE]”, with English translations. Unlike the word pairs used for constructing the SVM gender subspace (Table~\ref{tab:embedding_pairs}), this list includes pronouns like \textit{hij} and \textit{zij}, which are unambiguous in sentence context. Adjectives like \textit{mannelijk/vrouwelijk} are excluded due to structural incompatibility, and titles such as \textit{prins}, \textit{baron}, etc., are omitted as they sound unnatural in profession-based sentences.}
\label{tab:target_sentence_translated}
\end{table}

\FloatBarrier
\newpage

\section*{Appendix D: attribute terms} \label{app:d}

\begin{table}[H]
\centering
\renewcommand{\arraystretch}{0.9}
\footnotesize
\begin{tabular}{llll}
\textbf{Male Variant} & \textbf{Female Variant} & \textbf{\% Woman} & \textbf{English Translation} \\
\hline
apothekerassistent & apothekerassistente & 92.9\% & pharmacy assistant \\
praktijkassistent & praktijkassistente & 92.5\% & medical assistant \\
verzorger & verzorgster & 90.3\% & caregiver \\
schoonheidsspecialist & schoonheidsspecialiste & 90.2\% & beautician \\
kapper & kapster & 90.2\% & hairdresser \\
verpleger & verpleegster & 87.0\% & nurse \\
onderwijzer & onderwijzeres & 86.7\% & educator \\
psycholoog & psychologe & 79.6\% & psychologist \\
socioloog & sociologe & 79.6\% & sociologist \\
maatschappelijk werker & maatschappelijk werkster & 76.5\% & social worker \\
boekhouder & boekhoudster & 75.3\% & accountant \\
schoonmaker & schoonmaakster & 75.2\% & cleaner \\
fysiotherapeut & fysiotherapeute & 74.5\% & physiotherapist \\
laborant & laborante & 73.1\% & lab technician \\
sociaal werker & sociaal werkster & 73.8\% & social worker \\
administratief medewerker & administratief medewerkster & 72.4\% & administrative assistant \\
docent & docente & 66.1\% & teacher \\
verkoopmedewerker & verkoopmedewerkster & 65.2\% & sales assistant \\
verkoper & verkoopster & 65.2\% & salesperson \\
reisbegeleider & reisbegeleidster & 64.0\% & tour guide \\
hr-medewerker & hr-medewerkster & 63.4\% & HR employee \\
loopbaanadviseur & loopbaanadviseuse & 63.4\% & career advisor \\
loopbaanbegeleider & loopbaanbegeleidster & 63.4\% & career coach \\
barmedewerker & barmedewerkster & 62.8\% & bartender \\
bibliothecaris & bibliothecaresse & 62.5\% & librarian \\
receptionist & receptioniste & 67.2\% & receptionist \\
telefonist & telefoniste & 67.2\% & telephone operator \\
kunstenaar & kunstenares & 41.0\% & artist \\
bestuurder & bestuurster & 41.0\% & director \\
inspecteur & inspectrice & 40.0\% & inspector \\
politieagent & politieagente & 40.0\% & police officer \\
wetenschapper & wetenschapster & 38.0\% & scientist \\
bioloog & biologe & 38.0\% & biologist \\
architect & architecte & 35.7\% & architect \\
vertegenwoordiger & vertegenwoordigster & 31.0\% & representative \\
kok & kokkin & 30.3\% & cook \\
kleermaker & kleermaakster & 29.4\% & tailor \\
logistiek medewerker & logistiek medewerkster & 24.4\% & logistics employee \\
beveiligingsmedewerker & beveiligingsmedewerkster & 23.6\% & security guard \\
directeur & directrice & 23.1\% & director \\
chauffeur & chauffeuse & 13.0\% & chauffeur \\
productieleider & productieleidster & 9.9\% & production manager \\
schilder & schilderes & 6.7\% & painter \\
vrachtwagenchauffeur & vrachtwagenchauffeusse & 5.3\% & truck driver \\
constructiewerker & constructiewerkster & 3.2\% & construction worker \\
metaalbewerker & metaalbewerkster & 3.2\% & metalworker \\
\end{tabular}
\caption{Profession attributes in Dutch that exist in both male and female lexical forms, along with their gender distribution (\% women) and English translations. These attributes were used to create sentence pairs for evaluating stereotype alignment. The percentage of women employed in each role was used to categorize professions as stereotypically male or female, based on real-world gender distributions.
}
\label{tab:attributes-both-forms}
\end{table}

\FloatBarrier
\newpage

\section*{Appendix E: Statistical Significance Tests} \label{app:e}

To evaluate the \textbf{significance of differences in model accuracy} across conditions, we conducted two-proportion $z$-tests. The tables below summarize the comparisons, corresponding $z$-statistics, and $p$-values.

\begin{table}[h]
\centering
\renewcommand{\arraystretch}{1.2}
\begin{tabularx}{\textwidth}{>{\RaggedRight\arraybackslash}Xcc}
\toprule
\textbf{Comparison} & \textbf{$z$-statistic} & \textbf{$p$-value} \\
\midrule
\textbf{General Stereotype Effect} & & \\
Pro-tereotypical vs Anti-stereotypical (All) & 35.5452 & $<$0.0001 \\
\midrule
\textbf{Within-Gender Comparisons} & & \\
Female: Anti-stereotypical vs Pro-tereotypical & -23.4358 & $<$0.0001 \\
Male: Anti-stereotypical vs Pro-stereotypical & -25.4602 & $<$0.0001 \\
\midrule
\textbf{Cross-Gender Comparisons (Same Context)} & & \\
Female Anti-stereotypical vs Male Anti-stereotypical & 3.3015 & 0.0010 \\
Female Pro-stereotypical vs Male Pro-stereotypical & 2.0284 & 0.0425 \\
\midrule
\textbf{Effects of female-marked attributes (Female-suffix vs Neutral)} & & \\
Female Anti-stereotypical: Female-suffix vs Neutral & 24.0283 & $<$0.0001 \\
Female Pro-stereotypical: Female-suffix vs Neutral & 17.6232 & $<$0.0001 \\
\bottomrule
\end{tabularx}
\caption{Results of two-proportion $z$-tests comparing model accuracy across stereotype alignment, target gender, and the use of female-marked suffixes in professions versus neutral professions. Each row shows the $z$-statistic and corresponding $p$-value for a specific comparison. All effects are highly significant (p $<$ 0.0001), confirming robust performance differences across conditions.}
\label{tab:ztests}
\end{table}


\bibliographystyle{clin} 
\bibliography{bibliography}  

@article{BERT,
  author       = {Jacob Devlin and
                  Ming{-}Wei Chang and
                  Kenton Lee and
                  Kristina Toutanova},
  title        = {{BERT:} Pre-training of Deep Bidirectional Transformers for Language
                  Understanding},
  journal      = {CoRR},
  volume       = {abs/1810.04805},
  year         = {2018},
  url          = {http://arxiv.org/abs/1810.04805},
  eprinttype    = {arXiv},
  eprint       = {1810.04805},
  bibsource    = {dblp computer science bibliography, https://dblp.org}
}

@inproceedings{peters-etal-2018-ELMo,
    title = "Deep Contextualized Word Representations",
    author = "Peters, Matthew E.  and
      Neumann, Mark  and
      Iyyer, Mohit  and
      Gardner, Matt  and
      Clark, Christopher  and
      Lee, Kenton  and
      Zettlemoyer, Luke",
    editor = "Walker, Marilyn  and
      Ji, Heng  and
      Stent, Amanda",
    booktitle = "Proceedings of the 2018 Conference of the North {A}merican Chapter of the Association for Computational Linguistics: Human Language Technologies, Volume 1 (Long Papers)",
    month = jun,
    year = "2018",
    address = "New Orleans, Louisiana",
    publisher = "Association for Computational Linguistics",
    url = "https://aclanthology.org/N18-1202/",
    doi = "10.18653/v1/N18-1202",
    pages = "2227--2237"
}

@article{Tan&Celis,
  author       = {Yi Chern Tan and
                  L. Elisa Celis},
  title        = {Assessing Social and Intersectional Biases in Contextualized Word
                  Representations},
  journal      = {CoRR},
  volume       = {abs/1911.01485},
  year         = {2019},
  url          = {http://arxiv.org/abs/1911.01485},
  eprinttype    = {arXiv},
  eprint       = {1911.01485},
  bibsource    = {dblp computer science bibliography, https://dblp.org}
}

@inproceedings{StochasticParrots,
author = {Bender, Emily M. and Gebru, Timnit and McMillan-Major, Angelina and Shmitchell, Shmargaret},
title = {On the Dangers of Stochastic Parrots: Can Language Models Be Too Big?},
year = {2021},
isbn = {9781450383097},
publisher = {Association for Computing Machinery},
address = {New York, NY, USA},
url = {https://doi.org/10.1145/3442188.3445922},
doi = {10.1145/3442188.3445922},
booktitle = {Proceedings of the 2021 ACM Conference on Fairness, Accountability, and Transparency},
pages = {610–623},
numpages = {14},
location = {Virtual Event, Canada},
series = {FAccT '21}
}

@inproceedings{vanmassenhove-etal-2018-getting,
    title = "Getting Gender Right in Neural Machine Translation",
    author = "Vanmassenhove, Eva  and
      Hardmeier, Christian  and
      Way, Andy",
    editor = "Riloff, Ellen  and
      Chiang, David  and
      Hockenmaier, Julia  and
      Tsujii, Jun{'}ichi",
    booktitle = "Proceedings of the 2018 Conference on Empirical Methods in Natural Language Processing",
    month = oct # "-" # nov,
    year = "2018",
    address = "Brussels, Belgium",
    publisher = "Association for Computational Linguistics",
    url = "https://aclanthology.org/D18-1334/",
    doi = "10.18653/v1/D18-1334",
    pages = "3003--3008"
}

@inproceedings{saunders-byrne-2020-reducing,
    title = "Reducing Gender Bias in Neural Machine Translation as a Domain Adaptation Problem",
    author = "Saunders, Danielle  and
      Byrne, Bill",
    editor = "Jurafsky, Dan  and
      Chai, Joyce  and
      Schluter, Natalie  and
      Tetreault, Joel",
    booktitle = "Proceedings of the 58th Annual Meeting of the Association for Computational Linguistics",
    month = jul,
    year = "2020",
    address = "Online",
    publisher = "Association for Computational Linguistics",
    url = "https://aclanthology.org/2020.acl-main.690/",
    doi = "10.18653/v1/2020.acl-main.690",
    pages = "7724--7736"
}

@article{warner2024modernbert,
  title     = {Smarter, Better, Faster, Longer: A Modern Bidirectional Encoder for Fast, Memory Efficient, and Long Context Finetuning and Inference},
  author    = {Warner, Benjamin and Chaffin, Antoine and Clavi{\'e}, Benjamin and Weller, Orion and Hallstr{\"o}m, Oskar and Taghadouini, Said and Gallagher, Alexis and Biswas, Raja and Ladhak, Faisal and Aarsen, Tom and Cooper, Nathan and Adams, Griffin and Howard, Jeremy and Poli, Iacopo},
  journal   = {arXiv preprint arXiv:2412.13663},
  year      = {2024}
}

@article{boizard2025eurobert,
  title     = {EuroBERT: Scaling Multilingual Encoders for European Languages},
  author    = {Boizard, Nicolas and Gisserot-Boukhlef, Hippolyte and Alves, Duarte M. and Martins, Andr{\'e} and Hammal, Ayoub and Corro, Caio and Hudelot, C{\'e}line and Malherbe, Emmanuel and Malaboeuf, Etienne and Jourdan, Fanny and Hautreux, Gabriel and Alves, Jo{\~a}o and El-Haddad, Kevin and Faysse, Manuel and Peyrard, Maxime and Guerreiro, Nuno M. and Fernandes, Patrick and Rei, Ricardo and Colombo, Pierre},
  journal   = {arXiv preprint arXiv:2503.05500},
  year      = {2025}
}

@article{lebreton2025neobert,
  title     = {NeoBERT: A Next-Generation BERT},
  author    = {Le Breton, Lola and Fournier, Quentin and El Mezouar, Mariam and Chandar, Sarath},
  journal   = {Transactions on Machine Learning Research},
  year      = {2025}
}

@inproceedings{sap-etal-2019-risk,
    title = "The Risk of Racial Bias in Hate Speech Detection",
    author = "Sap, Maarten  and
      Card, Dallas  and
      Gabriel, Saadia  and
      Choi, Yejin  and
      Smith, Noah A.",
    editor = "Korhonen, Anna  and
      Traum, David  and
      M{\`a}rquez, Llu{\'i}s",
    booktitle = "Proceedings of the 57th Annual Meeting of the Association for Computational Linguistics",
    month = jul,
    year = "2019",
    address = "Florence, Italy",
    publisher = "Association for Computational Linguistics",
    url = "https://aclanthology.org/P19-1163/",
    doi = "10.18653/v1/P19-1163",
    pages = "1668--1678"
}

@inproceedings{fleisig-etal-2023-fairprism,
    title = "{F}air{P}rism: Evaluating Fairness-Related Harms in Text Generation",
    author = "Fleisig, Eve  and
      Amstutz, Aubrie  and
      Atalla, Chad  and
      Blodgett, Su Lin  and
      Daum{\'e} III, Hal  and
      Olteanu, Alexandra  and
      Sheng, Emily  and
      Vann, Dan  and
      Wallach, Hanna",
    editor = "Rogers, Anna  and
      Boyd-Graber, Jordan  and
      Okazaki, Naoaki",
    booktitle = "Proceedings of the 61st Annual Meeting of the Association for Computational Linguistics (Volume 1: Long Papers)",
    month = jul,
    year = "2023",
    address = "Toronto, Canada",
    publisher = "Association for Computational Linguistics",
    url = "https://aclanthology.org/2023.acl-long.343/",
    doi = "10.18653/v1/2023.acl-long.343",
    pages = "6231--6251"
}

@inproceedings{sheng-etal-2019-woman,
    title = "The Woman Worked as a Babysitter: On Biases in Language Generation",
    author = "Sheng, Emily  and
      Chang, Kai-Wei  and
      Natarajan, Premkumar  and
      Peng, Nanyun",
    editor = "Inui, Kentaro  and
      Jiang, Jing  and
      Ng, Vincent  and
      Wan, Xiaojun",
    booktitle = "Proceedings of the 2019 Conference on Empirical Methods in Natural Language Processing and the 9th International Joint Conference on Natural Language Processing (EMNLP-IJCNLP)",
    month = nov,
    year = "2019",
    address = "Hong Kong, China",
    publisher = "Association for Computational Linguistics",
    url = "https://aclanthology.org/D19-1339/",
    doi = "10.18653/v1/D19-1339",
    pages = "3407--3412"
}

@article{BolukbasiCZSK16a,
  author       = {Tolga Bolukbasi and
                  Kai{-}Wei Chang and
                  James Y. Zou and
                  Venkatesh Saligrama and
                  Adam Kalai},
  title        = {Man is to Computer Programmer as Woman is to Homemaker? Debiasing
                  Word Embeddings},
  journal      = {CoRR},
  volume       = {abs/1607.06520},
  year         = {2016},
  url          = {http://arxiv.org/abs/1607.06520},
  eprinttype    = {arXiv},
  eprint       = {1607.06520},
  bibsource    = {dblp computer science bibliography, https://dblp.org}
}

@inproceedings{ravfogel-etal-2020-null,
    title = "Null It Out: Guarding Protected Attributes by Iterative Nullspace Projection",
    author = "Ravfogel, Shauli  and
      Elazar, Yanai  and
      Gonen, Hila  and
      Twiton, Michael  and
      Goldberg, Yoav",
    editor = "Jurafsky, Dan  and
      Chai, Joyce  and
      Schluter, Natalie  and
      Tetreault, Joel",
    booktitle = "Proceedings of the 58th Annual Meeting of the Association for Computational Linguistics",
    month = jul,
    year = "2020",
    address = "Online",
    publisher = "Association for Computational Linguistics",
    url = "https://aclanthology.org/2020.acl-main.647/",
    doi = "10.18653/v1/2020.acl-main.647",
    pages = "7237--7256"
}

@misc{vanderwal2022birthbiascasestudy,
      title={The Birth of Bias: A case study on the evolution of gender bias in an English language model}, 
      author={Oskar van der Wal and Jaap Jumelet and Katrin Schulz and Willem Zuidema},
      year={2022},
      eprint={2207.10245},
      archivePrefix={arXiv},
      primaryClass={cs.CL},
      url={https://arxiv.org/abs/2207.10245}, 
}

@article{caliskan2017semantics,
  title={Semantics derived automatically from language corpora contain human-like biases},
  author={Caliskan, Aylin and Bryson, Joanna J and Narayanan, Arvind},
  journal={Science},
  volume={356},
  number={6334},
  pages={183--186},
  year={2017},
  publisher={American Association for the Advancement of Science}
}

@inproceedings{may-etal-2019-measuring,
    title = "On Measuring Social Biases in Sentence Encoders",
    author = "May, Chandler  and
      Wang, Alex  and
      Bordia, Shikha  and
      Bowman, Samuel R.  and
      Rudinger, Rachel",
    editor = "Burstein, Jill  and
      Doran, Christy  and
      Solorio, Thamar",
    booktitle = "Proceedings of the 2019 Conference of the North {A}merican Chapter of the Association for Computational Linguistics: Human Language Technologies, Volume 1 (Long and Short Papers)",
    month = jun,
    year = "2019",
    address = "Minneapolis, Minnesota",
    publisher = "Association for Computational Linguistics",
    url = "https://aclanthology.org/N19-1063/",
    doi = "10.18653/v1/N19-1063",
    pages = "622--628"
}

@inproceedings{zhao-etal-2018-gender,
    title = "Gender Bias in Coreference Resolution: Evaluation and Debiasing Methods",
    author = "Zhao, Jieyu  and
      Wang, Tianlu  and
      Yatskar, Mark  and
      Ordonez, Vicente  and
      Chang, Kai-Wei",
    editor = "Walker, Marilyn  and
      Ji, Heng  and
      Stent, Amanda",
    booktitle = "Proceedings of the 2018 Conference of the North {A}merican Chapter of the Association for Computational Linguistics: Human Language Technologies, Volume 2 (Short Papers)",
    month = jun,
    year = "2018",
    address = "New Orleans, Louisiana",
    publisher = "Association for Computational Linguistics",
    url = "https://aclanthology.org/N18-2003/",
    doi = "10.18653/v1/N18-2003",
    pages = "15--20"
}

@inproceedings{rudinger-etal-2018-gender,
    title = "Gender Bias in Coreference Resolution",
    author = "Rudinger, Rachel  and
      Naradowsky, Jason  and
      Leonard, Brian  and
      Van Durme, Benjamin",
    editor = "Walker, Marilyn  and
      Ji, Heng  and
      Stent, Amanda",
    booktitle = "Proceedings of the 2018 Conference of the North {A}merican Chapter of the Association for Computational Linguistics: Human Language Technologies, Volume 2 (Short Papers)",
    month = jun,
    year = "2018",
    address = "New Orleans, Louisiana",
    publisher = "Association for Computational Linguistics",
    url = "https://aclanthology.org/N18-2002/",
    doi = "10.18653/v1/N18-2002",
    pages = "8--14"
}

@inproceedings{bartl-etal-2020-unmasking,
    title = "Unmasking Contextual Stereotypes: Measuring and Mitigating {BERT}`s Gender Bias",
    author = "Bartl, Marion  and
      Nissim, Malvina  and
      Gatt, Albert",
    editor = "Costa-juss{\`a}, Marta R.  and
      Hardmeier, Christian  and
      Radford, Will  and
      Webster, Kellie",
    booktitle = "Proceedings of the Second Workshop on Gender Bias in Natural Language Processing",
    month = dec,
    year = "2020",
    address = "Barcelona, Spain (Online)",
    publisher = "Association for Computational Linguistics",
    url = "https://aclanthology.org/2020.gebnlp-1.1/",
    pages = "1--16"
}

@inproceedings{kurita-etal-2019-measuring,
    title = "Measuring Bias in Contextualized Word Representations",
    author = "Kurita, Keita  and
      Vyas, Nidhi  and
      Pareek, Ayush  and
      Black, Alan W  and
      Tsvetkov, Yulia",
    editor = "Costa-juss{\`a}, Marta R.  and
      Hardmeier, Christian  and
      Radford, Will  and
      Webster, Kellie",
    booktitle = "Proceedings of the First Workshop on Gender Bias in Natural Language Processing",
    month = aug,
    year = "2019",
    address = "Florence, Italy",
    publisher = "Association for Computational Linguistics",
    url = "https://aclanthology.org/W19-3823/",
    doi = "10.18653/v1/W19-3823",
    pages = "166--172"
}

@inproceedings{nangia-etal-2020-crows,
    title = "{C}row{S}-Pairs: A Challenge Dataset for Measuring Social Biases in Masked Language Models",
    author = "Nangia, Nikita  and
      Vania, Clara  and
      Bhalerao, Rasika  and
      Bowman, Samuel R.",
    editor = "Webber, Bonnie  and
      Cohn, Trevor  and
      He, Yulan  and
      Liu, Yang",
    booktitle = "Proceedings of the 2020 Conference on Empirical Methods in Natural Language Processing (EMNLP)",
    month = nov,
    year = "2020",
    address = "Online",
    publisher = "Association for Computational Linguistics",
    url = "https://aclanthology.org/2020.emnlp-main.154/",
    doi = "10.18653/v1/2020.emnlp-main.154",
    pages = "1953--1967"
}

@inproceedings{nadeem-etal-2021-stereoset,
    title = "{S}tereo{S}et: Measuring stereotypical bias in pretrained language models",
    author = "Nadeem, Moin  and
      Bethke, Anna  and
      Reddy, Siva",
    editor = "Zong, Chengqing  and
      Xia, Fei  and
      Li, Wenjie  and
      Navigli, Roberto",
    booktitle = "Proceedings of the 59th Annual Meeting of the Association for Computational Linguistics and the 11th International Joint Conference on Natural Language Processing (Volume 1: Long Papers)",
    month = aug,
    year = "2021",
    address = "Online",
    publisher = "Association for Computational Linguistics",
    url = "https://aclanthology.org/2021.acl-long.416/",
    doi = "10.18653/v1/2021.acl-long.416",
    pages = "5356--5371"
}

@inproceedings{salazar-etal-2020-masked,
    title = "Masked Language Model Scoring",
    author = "Salazar, Julian  and
      Liang, Davis  and
      Nguyen, Toan Q.  and
      Kirchhoff, Katrin",
    editor = "Jurafsky, Dan  and
      Chai, Joyce  and
      Schluter, Natalie  and
      Tetreault, Joel",
    booktitle = "Proceedings of the 58th Annual Meeting of the Association for Computational Linguistics",
    month = jul,
    year = "2020",
    address = "Online",
    publisher = "Association for Computational Linguistics",
    url = "https://aclanthology.org/2020.acl-main.240/",
    doi = "10.18653/v1/2020.acl-main.240",
    pages = "2699--2712"
}

@inproceedings{gonen-etal-2019-grammatical,
    title = "How Does Grammatical Gender Affect Noun Representations in Gender-Marking Languages?",
    author = "Gonen, Hila  and
      Kementchedjhieva, Yova  and
      Goldberg, Yoav",
    editor = "Bansal, Mohit  and
      Villavicencio, Aline",
    booktitle = "Proceedings of the 23rd Conference on Computational Natural Language Learning (CoNLL)",
    month = nov,
    year = "2019",
    address = "Hong Kong, China",
    publisher = "Association for Computational Linguistics",
    url = "https://aclanthology.org/K19-1043/",
    doi = "10.18653/v1/K19-1043",
    pages = "463--471"
}

@inproceedings{chavez-mulsa-spanakis-2020-evaluating,
    title = "Evaluating Bias In {D}utch Word Embeddings",
    author = "Ch{\'a}vez Mulsa, Rodrigo Alejandro  and
      Spanakis, Gerasimos",
    editor = "Costa-juss{\`a}, Marta R.  and
      Hardmeier, Christian  and
      Radford, Will  and
      Webster, Kellie",
    booktitle = "Proceedings of the Second Workshop on Gender Bias in Natural Language Processing",
    month = dec,
    year = "2020",
    address = "Barcelona, Spain (Online)",
    publisher = "Association for Computational Linguistics",
    url = "https://aclanthology.org/2020.gebnlp-1.6/",
    pages = "56--71"
}

@inproceedings{zhou-etal-2019-examining,
    title = "Examining Gender Bias in Languages with Grammatical Gender",
    author = "Zhou, Pei  and
      Shi, Weijia  and
      Zhao, Jieyu  and
      Huang, Kuan-Hao  and
      Chen, Muhao  and
      Cotterell, Ryan  and
      Chang, Kai-Wei",
    editor = "Inui, Kentaro  and
      Jiang, Jing  and
      Ng, Vincent  and
      Wan, Xiaojun",
    booktitle = "Proceedings of the 2019 Conference on Empirical Methods in Natural Language Processing and the 9th International Joint Conference on Natural Language Processing (EMNLP-IJCNLP)",
    month = nov,
    year = "2019",
    address = "Hong Kong, China",
    publisher = "Association for Computational Linguistics",
    url = "https://aclanthology.org/D19-1531/",
    doi = "10.18653/v1/D19-1531",
    pages = "5276--5284"
}

@inproceedings{neveol-etal-2022-french,
    title = "{F}rench {C}row{S}-Pairs: Extending a challenge dataset for measuring social bias in masked language models to a language other than {E}nglish",
    author = {N{\'e}v{\'e}ol, Aur{\'e}lie  and
      Dupont, Yoann  and
      Bezan{\c{c}}on, Julien  and
      Fort, Kar{\"e}n},
    editor = "Muresan, Smaranda  and
      Nakov, Preslav  and
      Villavicencio, Aline",
    booktitle = "Proceedings of the 60th Annual Meeting of the Association for Computational Linguistics (Volume 1: Long Papers)",
    month = may,
    year = "2022",
    address = "Dublin, Ireland",
    publisher = "Association for Computational Linguistics",
    url = "https://aclanthology.org/2022.acl-long.583/",
    doi = "10.18653/v1/2022.acl-long.583",
    pages = "8521--8531"
}

@inproceedings{kaneko-bollegala-2021-debiasing,
    title = "Debiasing Pre-trained Contextualised Embeddings",
    author = "Kaneko, Masahiro  and
      Bollegala, Danushka",
    editor = "Merlo, Paola  and
      Tiedemann, Jorg  and
      Tsarfaty, Reut",
    booktitle = "Proceedings of the 16th Conference of the European Chapter of the Association for Computational Linguistics: Main Volume",
    month = apr,
    year = "2021",
    address = "Online",
    publisher = "Association for Computational Linguistics",
    url = "https://aclanthology.org/2021.eacl-main.107/",
    doi = "10.18653/v1/2021.eacl-main.107",
    pages = "1256--1266"
}

@misc{cbs2024beroepen,
  author       = {{Centraal Bureau voor de Statistiek (CBS)}},
  title        = {{Beroepsbevolking; beroep, beroepensectie, geslacht, persoonskenmerken, 2024}},
  year         = {2024},
  url          = {\url{https://opendata.cbs.nl/#/CBS/nl/dataset/85276NED/table}},
  note         = {Accessed: 2025-04-30}
}

@inproceedings{arps-etal-2024-multilingual,
    title = "Multilingual Nonce Dependency Treebanks: Understanding how Language Models Represent and Process Syntactic Structure",
    author = "Arps, David  and
      Kallmeyer, Laura  and
      Samih, Younes  and
      Sajjad, Hassan",
    editor = "Duh, Kevin  and
      Gomez, Helena  and
      Bethard, Steven",
    booktitle = "Proceedings of the 2024 Conference of the North American Chapter of the Association for Computational Linguistics: Human Language Technologies (Volume 1: Long Papers)",
    month = jun,
    year = "2024",
    address = "Mexico City, Mexico",
    publisher = "Association for Computational Linguistics",
    url = "https://aclanthology.org/2024.naacl-long.433/",
    doi = "10.18653/v1/2024.naacl-long.433",
    pages = "7822--7844"
}

@inproceedings{manna2025payingattentionherinvestigating,
    title = "Are We Paying Attention to Her? Investigating Gender Disambiguation and Attention in Machine Translation",
    author = "Manna, Chiara  and
      Alishahi, Afra  and
      Blain, Fr{\'e}d{\'e}ric  and
      Vanmassenhove, Eva",
    booktitle = "Proceedings of the 3rd Workshop on Gender-Inclusive Translation Technologies (GITT 2025)",
    month = jun,
    year = "2025",
    address = "Geneva, Switzerland",
    publisher = "European Association for Machine Translation",
    url = "https://aclanthology.org/2025.gitt-1.1/",
    pages = "1--16",
}

@inproceedings{hewitt-manning-2019-structural,
    title = "{A} Structural Probe for Finding Syntax in Word Representations",
    author = "Hewitt, John  and
      Manning, Christopher D.",
    editor = "Burstein, Jill  and
      Doran, Christy  and
      Solorio, Thamar",
    booktitle = "Proceedings of the 2019 Conference of the North {A}merican Chapter of the Association for Computational Linguistics: Human Language Technologies, Volume 1 (Long and Short Papers)",
    month = jun,
    year = "2019",
    address = "Minneapolis, Minnesota",
    publisher = "Association for Computational Linguistics",
    url = "https://aclanthology.org/N19-1419/",
    doi = "10.18653/v1/N19-1419",
    pages = "4129--4138"
}

@book{bird2009natural,
  title     = {Natural Language Processing with Python: Analyzing Text with the Natural Language Toolkit},
  author    = {Bird, Steven and Klein, Ewan and Loper, Edward},
  year      = {2009},
  publisher = {O'Reilly Media, Inc.},
  isbn      = {9780596516499},
  url       = {https://www.nltk.org/book/}
}

@article{Vaswanietal,
  author       = {Ashish Vaswani and
                  Noam Shazeer and
                  Niki Parmar and
                  Jakob Uszkoreit and
                  Llion Jones and
                  Aidan N. Gomez and
                  Lukasz Kaiser and
                  Illia Polosukhin},
  title        = {Attention Is All You Need},
  journal      = {CoRR},
  volume       = {abs/1706.03762},
  year         = {2017},
  url          = {http://arxiv.org/abs/1706.03762},
  eprinttype    = {arXiv},
  eprint       = {1706.03762},
  bibsource    = {dblp computer science bibliography, https://dblp.org}
}

@article{SVMs,
author = {Cortes, Corinna and Vapnik, Vladimir},
title = {Support-Vector Networks},
year = {1995},
issue_date = {Sept. 1995},
publisher = {Kluwer Academic Publishers},
address = {USA},
volume = {20},
number = {3},
issn = {0885-6125},
url = {https://doi.org/10.1023/A:1022627411411},
doi = {10.1023/A:1022627411411},
journal = {Mach. Learn.},
month = sep,
pages = {273–297},
numpages = {25}
}

@article{oostdijk2013construction,
  title={The Construction of a 500-Million-Word Reference Corpus of Contemporary Written Dutch},
  author={Oostdijk, Nelleke and Reynaert, Martin and Hoste, Veronique and Schuurman, Ineke},
  journal={Essential Speech and Language Technology for Dutch: Results by the STEVIN-programme},
  pages={219},
  year={2013},
  publisher={Springer Science \& Business Media}
}

@article{mortelmans2008zij,
  title={Zij is een powerfeministe. Nog eens functie-en rolbenamingen in het Nederlands vanuit contastief perspectief},
  author={Mortelmans, Tanja},
  journal={Tijdschrift voor genderstudies},
  volume={11},
  number={1},
  year={2008}
}

@article{gerritsen2002towards,
  title={Towards a more gender-fair usage in Netherlands Dutch},
  author={Gerritsen, Marinel},
  journal={Gender Across Languages: The linguistic representation of women and men},
  volume={2},
  pages={81},
  year={2002},
  publisher={John Benjamins Publishing}
}

@article{romein1975over,
  title={Over taal en seks, seksisme en emancipatie},
  author={Romein-Verschoor, Annie},
  journal={De Gids},
  volume={138},
  number={1/2},
  pages={3--36},
  year={1975}
}

@mastersthesis{boudewijn2023masculinegeneric,
  author       = {Boudewijn, Julia},
  title        = {Alternatives to the Masculine Generic in Two Countries: 
                  A Comparative Study of Mental Representations of Gender 
                  in the Netherlands and Belgium},
  school       = {Leiden University},
  address      = {Leiden, The Netherlands},
  year         = {2023},
  type         = {Master's thesis},
  note         = {MSc Applied Cognitive Psychology, Supervisor: Juan Olvido Perea-García},
}

@article{vanmassenhove2025losing,
  title={Losing our Tail--Again: On (Un) Natural Selection And Multilingual Large Language Models},
  author={Vanmassenhove, Eva},
  journal={arXiv preprint arXiv:2507.03933},
  year={2025}
}

@inproceedings{vanmassenhove2019lost,
  title={Lost in Translation: Loss and Decay of Linguistic Richness in Machine Translation},
  author={Vanmassenhove, Eva and Shterionov, Dimitar and Way, Andy},
  booktitle={Proceedings of Machine Translation Summit XVII: Research Track},
  pages={222--232},
  year={2019}
}

@inproceedings{vanmassenhove2021machine,
  title={Machine Translationese: Effects of Algorithmic Bias on Linguistic Complexity in Machine Translation},
  author={Vanmassenhove, Eva and Shterionov, Dimitar and Gwilliam, Matthew},
  booktitle={Proceedings of the 16th Conference of the European Chapter of the Association for Computational Linguistics: Main Volume},
  pages={2203--2213},
  year={2021}
}

@article{mccoy2023embers,
  title={Embers of autoregression: Understanding large language models through the problem they are trained to solve},
  author={McCoy, R Thomas and Yao, Shunyu and Friedman, Dan and Hardy, Matthew and Griffiths, Thomas L},
  journal={arXiv preprint arXiv:2309.13638},
  year={2023}
}

@article{savoldi2025decade,
  title={A decade of gender bias in machine translation},
  author={Savoldi, Beatrice and Bastings, Jasmijn and Bentivogli, Luisa and Vanmassenhove, Eva},
  journal={Patterns},
  volume={6},
  number={6},
  year={2025},
  publisher={Elsevier}
}

@article{vervecken2013changing,
  title={Changing (S) expectations: How gender fair job descriptions impact children's perceptions and interest regarding traditionally male occupations},
  author={Vervecken, Dries and Hannover, Bettina and Wolter, Ilka},
  journal={Journal of Vocational Behavior},
  volume={82},
  number={3},
  pages={208--220},
  year={2013},
  publisher={Elsevier}
}

@article{vervecken2015yes,
  title={Yes i can!Effects of gender fair job descriptions on children’s perceptions of job status, job difficulty, and vocational self-efficacy},
  author={Vervecken, Dries and Hannover, Bettina},
  journal={Social Psychology},
  year={2015},
  publisher={Hogrefe Publishing}
}

@article{liu2019roberta,
  title={Roberta: A robustly optimized bert pretraining approach},
  author={Liu, Yinhan and Ott, Myle and Goyal, Naman and Du, Jingfei and Joshi, Mandar and Chen, Danqi and Levy, Omer and Lewis, Mike and Zettlemoyer, Luke and Stoyanov, Veselin},
  journal={arXiv preprint arXiv:1907.11692},
  year={2019}
}

@incollection{van2026gender,
  title={Gender-inclusive language in Dutch},
  author={Van Hoof, Sarah and Decock, Sofie},
  booktitle={Gender-Inclusive Language. Findings from 14 Languages and Open Research Questions},
  pages={41--67},
  year={2026},
  publisher={De Gruyter Brill}
}

@article{steurs2021hoe,
  title={Hoe automatische vertaling de genderbias van AI (Artificial Intelligence) verraadt},
  author={Steurs, Frieda and Vandeghinste, Vincent and Heylen, Kris},
  year={2021},
  publisher={Sterck De Vreese}
}

@inproceedings{fort-etal-2024-stereotypical,
    title = "Your Stereotypical Mileage May Vary: Practical Challenges of Evaluating Biases in Multiple Languages and Cultural Contexts",
    author = "Fort, Karen  and
      Alonso Alemany, Laura  and
      Benotti, Luciana  and
      Bezan{\c{c}}on, Julien  and
      Borg, Claudia  and
      Borg, Marthese  and
      Chen, Yongjian  and
      Ducel, Fanny  and
      Dupont, Yoann  and
      Ivetta, Guido  and
      Li, Zhijian  and
      Mieskes, Margot  and
      Naguib, Marco  and
      Qian, Yuyan  and
      Radaelli, Matteo  and
      Schmeisser-Nieto, Wolfgang S.  and
      Raimundo Schulz, Emma  and
      Saci, Thiziri  and
      Saidi, Sarah  and
      Torroba Marchante, Javier  and
      Xie, Shilin  and
      Zanotto, Sergio E.  and
      N{\'e}v{\'e}ol, Aur{\'e}lie",
    editor = "Calzolari, Nicoletta  and
      Kan, Min-Yen  and
      Hoste, Veronique  and
      Lenci, Alessandro  and
      Sakti, Sakriani  and
      Xue, Nianwen",
    booktitle = "Proceedings of the 2024 Joint International Conference on Computational Linguistics, Language Resources and Evaluation (LREC-COLING 2024)",
    month = may,
    year = "2024",
    address = "Torino, Italia",
    publisher = "ELRA and ICCL",
    url = "https://aclanthology.org/2024.lrec-main.1545/",
    pages = "17764--17769"
}

@inproceedings{liang-etal-2020-monolingual,
    title = "Monolingual and Multilingual Reduction of Gender Bias in Contextualized Representations",
    author = {Liang, Sheng  and
      Dufter, Philipp  and
      Sch{\"u}tze, Hinrich},
    editor = "Scott, Donia  and
      Bel, Nuria  and
      Zong, Chengqing",
    booktitle = "Proceedings of the 28th International Conference on Computational Linguistics",
    month = dec,
    year = "2020",
    address = "Barcelona, Spain (Online)",
    publisher = "International Committee on Computational Linguistics",
    url = "https://aclanthology.org/2020.coling-main.446/",
    doi = "10.18653/v1/2020.coling-main.446",
    pages = "5082--5093"
}

@inproceedings{jawahar-etal-2019-bert,
    title = "What Does {BERT} Learn about the Structure of Language?",
    author = "Jawahar, Ganesh  and
      Sagot, Beno{\^i}t  and
      Seddah, Djam{\'e}",
    editor = "Korhonen, Anna  and
      Traum, David  and
      M{\`a}rquez, Llu{\'i}s",
    booktitle = "Proceedings of the 57th Annual Meeting of the Association for Computational Linguistics",
    month = jul,
    year = "2019",
    address = "Florence, Italy",
    publisher = "Association for Computational Linguistics",
    url = "https://aclanthology.org/P19-1356/",
    doi = "10.18653/v1/P19-1356",
    pages = "3651--3657"}

@inproceedings{tenney-etal-2019-bert,
    title = "{BERT} Rediscovers the Classical {NLP} Pipeline",
    author = "Tenney, Ian  and
      Das, Dipanjan  and
      Pavlick, Ellie",
    editor = "Korhonen, Anna  and
      Traum, David  and
      M{\`a}rquez, Llu{\'i}s",
    booktitle = "Proceedings of the 57th Annual Meeting of the Association for Computational Linguistics",
    month = jul,
    year = "2019",
    address = "Florence, Italy",
    publisher = "Association for Computational Linguistics",
    url = "https://aclanthology.org/P19-1452/",
    doi = "10.18653/v1/P19-1452",
    pages = "4593--4601"
}

\end{document}